\documentclass[sigconf,9pt,screen]{acmart}

\AtBeginDocument{%
  }

\usepackage[noend]{algpseudocode}
\usepackage[ruled,linesnumbered,vlined]{algorithm2e}
\usepackage{graphicx}
\usepackage{textcomp}
\usepackage{xcolor}  
\usepackage{multicol}
\usepackage{multirow}

\usepackage{subfigure}
\usepackage{wrapfig}
\usepackage{bm}
\usepackage{array}
\usepackage{booktabs}
\usepackage{setspace}
\usepackage{makecell}
\usepackage{hyperref}
\usepackage{supertabular}
\usepackage{pifont}
\usepackage{mathrsfs}
\usepackage{xspace}
\usepackage{acronym}
\usepackage[abbreviations]{foreign}
\usepackage{url}
\usepackage{csquotes}
\usepackage{cleveref}
\usepackage{subcaption}
\usepackage{caption}
\usepackage{siunitx}
\usepackage{soul}
\usepackage{threeparttable}
\usepackage{enumitem}
\usepackage{tikz}
\newcommand*\circled[1]{\tikz[baseline=(char.base)]{
            \node[shape=circle,draw,inner sep=1.6pt] (char) {#1};}}

\usepackage{pifont}
\newcommand{\cmark}{\ding{51}}%
\newcommand{\xmark}{\ding{55}}%
\newcommand{\greencheck}{{\color{green}\cmark}}
\newcommand{\redcross}{{\color{red}\xmark}}

\definecolor{lightred}{rgb}{1, 0.8, 0.8}   
\definecolor{lightgreen}{rgb}{0.8, 1, 0.8} 

\usepackage{amsmath,amsfonts,bm}

\def\eqref#1{equation~\ref{#1}}

\def\1{\bm{1}}

\DeclareMathAlphabet{\mathsfit}{\encodingdefault}{\sfdefault}{m}{sl}
\SetMathAlphabet{\mathsfit}{bold}{\encodingdefault}{\sfdefault}{bx}{n}

\DeclareMathOperator*{\argmin}{arg\,min}

\newcommand{\bx}{\bm{x}}

\newcommand{\bs}{\bm{s}}
\newcommand{\btheta}{\bm{\theta}}

\newcommand{\mx}{\mathcal{X}}

\newcommand{\md}{D}
\newcommand{\ms}{S}
\newcommand{\mz}{Z}
\newcommand{\ml}{\mathcal{L}}
\newboolean{showcomments}
\setboolean{showcomments}{true}
\ifthenelse{\boolean{showcomments}}
{ \newcommand{\mynote}[3]{
		\fbox{\bfseries\sffamily\scriptsize#1}
		{\small$\blacktriangleright$\textsf{\emph{\color{#3}{#2}}}$\blacktriangleleft$}}
	\newcommand{\zzz}[1]{{\setlength{\fboxsep}{2pt}\fcolorbox{black}{yellow}{\textsf{\emph{#1}}}}\xspace}}
{ \newcommand{\mynote}[3]{}
	\newcommand{\zzz}[1]{}}

\definecolor{BlueViolet}{RGB}{138,43,226}

\acrodef{DL}{decentralized learning}
\acrodef{ML}{machine learning}
\acrodef{D-PSGD}{decentralized parallel stochastic gradient descent}
\acrodef{FL}{federated learning}
\acrodef{FI}{federated inference}
\acrodef{FU}{federated unlearning}
\acrodef{SGD}{stochastic gradient descent}
\acrodef{IID}{independent and identically distributed}
\acrodef{non-IID}{non independent and identically distributed}
\acrodef{RMSE}{root mean square error}
\acrodef{RMW}{random model walk}
\acrodef{GL}{gossip learning}
\acrodef{DWT}{discrete wavelet transform}
\acrodef{LAN}{local area network}
\acrodef{WAN}{wide area network}
\acrodef{NN}{neural network}
\acrodef{KD}{knowledge distillation}
\acrodef{DD}{dataset distillation}
\acrodef{GDPR}{General Data Protection Regulation}
\acrodef{SOTA}{state-of-the-art}
\acrodef{CCPA}{California Consumer Privacy Act}
\acrodef{MIA}{membership inference attack}
\acrodef{SGA}{stochastic gradient ascent}
\acrodef{SGD}{stochastic gradient descent}
\acrodef{MU}{machine unlearning}
\newcommand{\cifar}{CIFAR-10\xspace}
\newcommand{\mnist}{MNIST\xspace}
\newcommand{\svhn}{SVHN\xspace}
\newcommand{\fset}{F-Set\xspace}
\newcommand{\rset}{R-Set\xspace}
\newcommand{\retrainor}{\textsc{Retrain-Or}\xspace}
\newcommand{\sgaor}{\textsc{SGA-Or}\xspace}
\newcommand{\fump}{\textsc{FU-MP}\xspace}
\newcommand{\federaser}{\textsc{FedEraser}\xspace}

\newcommand{\su}{\textsc{S2U}\xspace}
\newcommand{\forgetset}{forget dataset\xspace}
\newcommand{\retainset}{retain dataset\xspace}

\newcommand{\fedavg}{\textsc{FedAvg}\xspace}
\newcommand{\sga}{\textsc{SGA}\xspace}
\newcommand{\sgd}{\textsc{SGD}\xspace}

\acmYear{2024}\copyrightyear{2024}
\acmBooktitle{24th International Middleware Conference (MIDDLEWARE '24), December 2--6, 2024, Hong Kong, Hong Kong}
\acmDOI{10.1145/3652892.3700764}

\setcopyright{rightsretained}
\copyrightyear{2024}

\acmConference[MIDDLEWARE '24]{24th International Middleware Conference}{December 2--6, 2024}{Hong Kong, Hong Kong}
\acmISBN{979-8-4007-0623-3/24/12}

\begin{document}

\newcommand{\sys}{\textsc{QuickDrop}\xspace}

\title{\sys: Efficient Federated Unlearning via Synthetic Data Generation}

\author{Akash Dhasade}
\affiliation{%
  \institution{EPFL}
  \country{Switzerland}
}

\author{Yaohong Ding}
\affiliation{%
  \institution{The Hong Kong Polytechnic University}
  \country{China}
}

\author{Song Guo}
\affiliation{%
  \institution{The Hong Kong University of Science and Technology}
  \country{China}
}

\author{Anne-Marie Kermarrec}
\affiliation{%
  \institution{EPFL}
  \country{Switzerland}
}

\author{Martijn de Vos}
\affiliation{%
  \institution{EPFL}
  \country{Switzerland}
}

\author{Leijie Wu}
\affiliation{%
  \institution{The Hong Kong Polytechnic University}
  \country{China}
}

\renewcommand{\shortauthors}{Dhasade et al.}

\begin{abstract}
Federated Unlearning (FU) aims to delete specific training data from an ML model trained using Federated Learning (FL).
However, existing FU methods suffer from inefficiencies due to the high costs associated with gradient recomputation and storage. 
This paper presents \sys, an original and efficient FU approach designed to overcome these limitations. 
During model training, each client uses \sys to generate a compact synthetic dataset, serving as a compressed representation of the gradient information utilized during training. 
This synthetic dataset facilitates fast gradient approximation, allowing rapid downstream unlearning at minimal storage cost. 
To unlearn some knowledge from the trained model, \sys clients execute stochastic gradient ascent with samples from the synthetic datasets instead of the training dataset.
The tiny volume of synthetic data significantly reduces computational overhead compared to conventional FU methods.
Evaluations with three standard datasets and five baselines show that, with comparable accuracy guarantees, \sys reduces the unlearning duration by $463\times$ compared to retraining the model from scratch and $65-218\times$ compared to FU baselines.
\sys supports both class- and client-level unlearning, multiple unlearning requests, and relearning of previously erased data.
\end{abstract}

\ccsdesc[300]{Information systems~Data management systems}
\ccsdesc[500]{Computing methodologies~Machine learning}

\keywords{Federated Unlearning, Machine Unlearning, Federated Learning, Privacy and Security, Dataset Distillation.}


\maketitle

\section{Introduction}
\label{introduction}

The vast amount of data produced by computing devices is increasingly used to train large-scale ML models that empower industrial processes and personal experiences~\cite{mahdavinejad2018machine}.
However, this data is often privacy sensitive or very large in volume, making it prohibitively expensive to upload it to a central server~\cite{bellet2018personalized,yang2019federated}.
To sidestep this issue, \ac{FL} is increasingly being applied to collaboratively train ML models in a privacy-preserving manner~\cite{mcmahan2017communication}.
\ac{FL} obviates the need to move the data to a central location by having participants only exchange model updates with a parameter server.
In each round of \ac{FL}, the parameter server aggregates all incoming trained models and then sends the aggregated global model back to participants.

Recent privacy regulations like the \ac{GDPR} and \ac{CCPA} grant data owners with the \enquote{right to be forgotten}~\cite{gdpr2018,ccpa2018}.
In the realm of ML, this requires organizations to be able to remove the influence of personal data on the trained model upon request~\cite{zhang2023review}.
This is called \emph{\ac{MU}}~\cite{bourtoule2021machine}.
For instance, hospitals that collaboratively trained a model using \ac{FL} might have to unlearn particular data samples in response to patient requests.
In \ac{FL}, beyond the ``right to be forgotten", removing data from the model proves essential for several other purposes. 
For instance, the ability to quickly eliminate outdated, manipulated, or erroneously included data enhances the security, responsiveness, and reliability of \ac{FL} systems~\cite{chundawat2023zero}.
However, the distributed nature of \ac{FL} and the inability of the parameter server to access training data directly makes \emph{\ac{FU}} a challenging task.

\begin{figure*}[t]
    \centering
    \includegraphics[width=.95\linewidth]{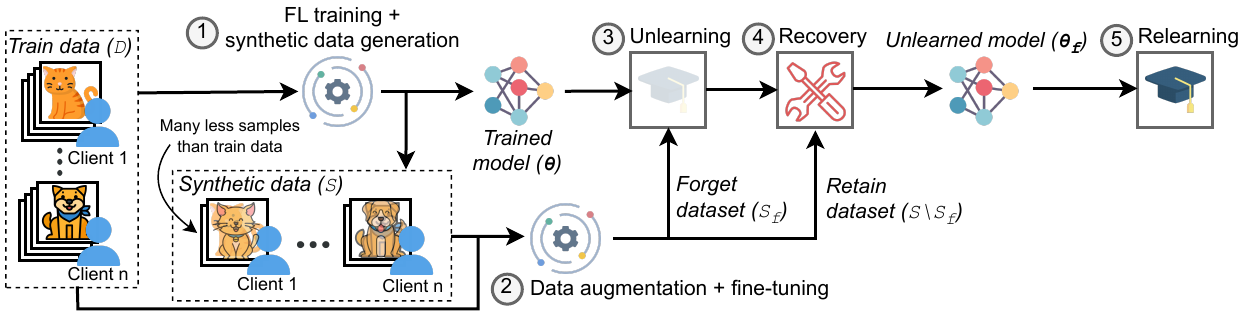}
    \caption{The workflow of \sys, our efficient federated unlearning method using synthetic data.} 
    \label{fig:quickdrop}
\end{figure*}

A naive method involves retraining the model from scratch, excluding target samples. 
This triggers many new training rounds, forcing clients to recompute gradients on the remaining data. 
Therefore, retraining is prohibitively expensive, with respect to the time and compute resources required.
Alternatively, some methods store gradients from original \ac{FL} training for \textit{downstream unlearning}~\cite{liu2021federaser}. 
However, the linear scaling of storage costs with the number of clients and \ac{FL} rounds leads to significant overhead. 
A more effective approach uses \textit{\ac{SGA}} on target samples~\cite{wu2022federated}, optimizing in the direction that maximizes the loss function. 
However, this approach updates the entire model, causing performance deterioration of the remaining samples.
A subsequent recovery phase is necessary, involving model retraining with the remaining samples for a few rounds. 
While \sga is more efficient than full retraining, it remains computationally demanding, requiring clients to recompute numerous gradients for unlearning and recovery. 
Thus, existing \ac{FU} approaches are inefficient and incur high costs for either gradient storage or recomputation.

To address this issue, we introduce \sys, a novel \ac{FU} approach that efficiently performs unlearning using \ac{SGA} and synthetic data. 
\sys circumvents the need to store or recompute expensive gradients by employing a technique of \textit{gradient matching}~\cite{zhao2021dataset} during \ac{FL} training. 
This process synthesizes a compact, client-specific dataset in situ, acting as a compressed representation of original gradients. 
This synthetic dataset facilitates \textit{fast gradient approximation} for downstream unlearning and significantly enhances efficiency.
Furthermore, this dataset is merely 1\% of the total volume of local datasets, resulting in minimal storage overhead.

The full workflow of \sys is shown in Figure~\ref{fig:quickdrop} and involves five main steps.
Initially, clients engage in regular FL training to collaboratively train a model (step \circled{1}). 
Simultaneously, each client generates a compact synthetic dataset through gradient matching, optimized for unlearning tasks. 
This synthetic dataset is then augmented with a few original samples and fine-tuned (step \circled{2}), which improves the model accuracy during recovery. 
Unlearning rounds (step \circled{3}) leverage \ac{SGA} on the local synthetic dataset, maintaining efficiency due to its small volume. 
Recovery rounds (step \circled{4}) also utilize the synthetic data, ensuring efficient downstream recovery compared to using original datasets. 
\sys also supports the efficient relearning of unlearned knowledge (step \circled{5}), again leveraging the synthetic data.
In summary, our approach overcomes the limitations of traditional methods, offering a streamlined and efficient solution for \ac{FU}.
%


\textbf{Contributions.} 
This paper makes the following contributions:
\begin{enumerate}
    \item We introduce \sys, a novel and efficient federated unlearning approach that generates and uses synthetic datasets to unlearn specific knowledge from a trained \ac{FL} model (\Cref{sec:quickdrop_design}).
    \item We formulate the problem of synthetic data generation for unlearning and show how such synthetic data can be generated in situ in \ac{FL} training by adapting the technique of gradient matching~\cite{zhao2021dataset} (\Cref{subsec:dd_formulation}).
    \item We implement and open-source \sys, and evaluate its unlearning performance in terms of efficiency and accuracy on three standard datasets and five \ac{SOTA} baselines (Section~\ref{sec:experiments}). Under comparable accuracy guarantees, we find that \sys reduces the duration of class unlearning by \textbf{463$\times$} compared to model retraining from scratch and \textbf{65-218$\times$} compared to other \ac{SOTA} \ac{FU} approaches.
\end{enumerate}
\section{Background and Problem Setup}
We first provide background on machine and federated unlearning in~\Cref{subsec:mu_and_fu}, then formulate the problem in~\Cref{subsec:preliminary}, and finally outline existing unlearning algorithms in~\Cref{subsec:existing_algos}.

\subsection{Machine and Federated Unlearning}
\label{subsec:mu_and_fu}
\textbf{\Acf{MU}} was first proposed by Cao and Yang~\cite{cao2015towards}.
Consider a set of trained weights $\btheta$ on a training dataset $D$.
The purpose of \ac{MU} is to remove the influence of a specific subset of $D_f \subset D$ on the model parameters $\btheta$.
The subset $D_f$ is generally referred to as the \emph{\forgetset}, and its complement $D \backslash D_f$ is called the \emph{\retainset}. 
Let the unlearning algorithm $\mathcal{U}$ be defined as $\btheta_f = \mathcal{U}(\btheta, D_f)$, where $\btheta_f$ is the unlearned model.
Unlearning thus aims to obtain a model $\btheta_f$ that is equivalent in performance to a model trained only on $D \backslash D_f$.
In other words, the unlearned model $\btheta_f$ should perform well on $D \backslash D_f$ while performing relatively less well on $D_f$.

\textbf{\Acf{FU}} is a \ac{MU} technique where knowledge is removed from a trained model in a distributed and collaborative manner.
Generally, \ac{FU} is more challenging than \ac{MU} for the following two reasons.
First, client data is only available on the client's side and cannot be moved to a central server.
This mandates active participation by clients to perform unlearning.
It also implies that the parameter server cannot conduct any fine-grained operations at the data level, rendering many existing \ac{MU} techniques inapplicable in \ac{FL} settings.
Second, when the original model is also trained in a distributed, collaborative manner, \eg, using \ac{FL}, the parameter server does not always have access to intermediate, granular training information produced by clients.
Some \ac{MU} techniques rely on recorded training information to carry out an unlearning operation \cite{graves2021amnesiac}.
In \ac{FU} settings, however, the parameter server might be unable to collect specific training information for unlearning, such as model updates per batch for each client.

\subsection{Problem Setup}
\label{subsec:preliminary}
We consider a \ac{FL} system containing $N$ clients (\eg, PCs or mobile devices), where each client $i \in N$ holds a local training dataset $D_i$.
Clients collaboratively train a global FL model $\btheta$ using an \ac{FL} algorithm (\eg \fedavg~\cite{mcmahan2017communication}). 
Once the global model $\btheta$ is trained, the parameter server may receive an unlearning request for the \forgetset $D_f$.
The characterization of $D_f$ defines the type of unlearning performed.
We distinguish between the following three types of unlearning:
\begin{itemize}
    \item \textbf{Class-level unlearning.} This type of unlearning erases the knowledge of a target class.
    Consequently, $D_f$ encompasses the entire data of a class. 
    Denoting by $D_i^c$ the data of class $c$ with the $i$-th client, we have $D_f := \cup_i D_i^c$, when the target class is $c$.
    The union is over all clients $i$ which possess samples of class $c$. 
    In essence, $D_f$ for class-level unlearning is distributed across clients.
    \item \textbf{Client-level unlearning.} This type of unlearning erases the knowledge of a target client, \eg, when exercising the right to be forgotten.
    Here, $D_f$ contains the data of a single client. 
    When the target client is $i$, $D_f := D_i$.
    In this case, $D_f$ is concentrated on a single client.
    \item \textbf{Sample-level unlearning.} This type of unlearning erases the knowledge of one or more samples.
    Here, $D_f$ contains arbitrary samples from one or more clients.
    Sample-level unlearning is the most general and difficult form of unlearning \cite{wu2022federated,romandini2024federated}.
\end{itemize}

\textbf{Relearning.} While the main objective of our work is to unlearn efficiently, it can be desirable to relearn the unlearned knowledge in certain situations, \eg, when a client revokes its right to be forgotten.
Where unlearning erases the knowledge in $D_f$ from the trained model, relearning aims to add this knowledge back.

We remark that class- and client-level unlearning are the two most common use cases in federated settings \cite{wu2022federated,romandini2024federated}; very few works address sample-level unlearning, even in the context of \ac{MU}.
In this work, we also specifically focus on class-level and client-level unlearning.
We further discuss this aspect in~\Cref{subsec:limitations}.

\subsection{Existing FU Algorithms and their Drawbacks}
\label{subsec:existing_algos}
We now discuss existing \ac{FU} approaches and their drawbacks before presenting the design of \sys in \Cref{sec:quickdrop_design}.

\textbf{Retraining from scratch.} A naive way to unlearn $D_f$ is to retrain the model from scratch while omitting samples from $D_f$. 
While this algorithm perfectly achieves the desired goal, complete retraining is prohibitively expensive as it initiates new \ac{FL} training rounds on $D \backslash D_f$.
Even executing a single unlearning request in such a way is highly compute- and time-intensive.
We refer to this algorithm as \retrainor \ie, as a retraining oracle due to its ideal achieved performance.

\textbf{Gradient calibration.} One way to speed up retraining from scratch is to reuse gradient information from the original training to avoid regenerating all gradients from scratch. 
However, these gradients must be adapted based on the \forgetset $D_f$ and \retainset $D \backslash D_f$, through a process referred to as \textit{gradient calibration}.  
Algorithms employing gradient calibration like FedEraser~\cite{liu2021federaser} trade the central server's storage for unlearned model's construction time by leveraging historical parameter updates from FL training.
However, the storage costs can grow quite large while the efficiency gains compared to retraining from scratch remain modest ($ \sim 4\times$).

\textbf{\Acf{SGA}.} Another \ac{FU} approach involves performing \ac{SGA} steps on $D_f$~\cite{wu2022federated}.
In each \ac{FL} round, clients having data in $D_f$ perform local \ac{SGA} steps while the parameter server aggregates the received updates.
However, \sga training introduces noise that affects the performance of remaining data.
This noise necessitates subsequent recovery rounds during which clients engage in regular \sgd training on the \retainset $D \backslash D_f$.
An unlearning request thus encompasses unlearning on $D_f$ and recovery on $D \backslash D_f$ with each request updating the model with the entire dataset.
Therefore, \sga remains inefficient when dealing with high dataset volumes or when executing many unlearning requests.
\Cref{algo:sga} provides the pseudo code for \sga.

\textbf{\su.} Inspired by the observation that the up- or down-scaling of model updates can substantially influence the global model, \su scales down the forgetting client's updates while scaling up the updates of remaining clients~\cite{gao2022verifi}.  
Its unlearning and recovery stages are integrated and conducted simultaneously, like \retrainor.
\su is specifically designed to only support client-level unlearning.

\textbf{Model Pruning.}
\fump~\cite{wang2022federated} uses model pruning by first measuring the class discrimination of channels in the model (\ie, the relevance of different classes on the model channel) and then prunes the most relevant channel of the target class to unlearn it.
While \fump is much more efficient than \retrainor, it only applies to class-level unlearning. 
Additionally, the pruned channels prevent relearning as pruning irreversibly modifies the model.

\begin{table}[t]
 \caption{Comparison of \ac{FU} approaches and \sys.}
\begin{threeparttable}
    \centering
    \scalebox{0.8}{
    \begin{tabular}{c|c|c|c|c|c}
        \toprule
        \emph{Algorithms} & \begin{tabular}[c]{@{}c@{}} Class- \\ unlearn. \end{tabular} & \begin{tabular}[c]{@{}c@{}} Client- \\ unlearn. \end{tabular} & Relearn. & \begin{tabular}[c]{@{}c@{}} Storage \\ Eff. \end{tabular}  & \begin{tabular}[c]{@{}c@{}} Computation \\ Eff. \end{tabular} \\
        \midrule
        \retrainor & \greencheck & \greencheck & \greencheck &  \greencheck &  \redcross~(very low) \\
        \federaser~\cite{liu2021federaser} & \greencheck & \greencheck & \greencheck & \redcross &  \redcross~(low) \\
        \su~\cite{gao2022verifi} & \redcross & \greencheck & \greencheck & 
        \greencheck &  \redcross~(low) \\
        \sga~\cite{wu2022federated} & \greencheck & \greencheck & \greencheck & \greencheck &  \redcross~(medium) \\
        \fump~\cite{wang2022federated} & \greencheck & \redcross & \redcross & 
        \greencheck &  \redcross~(medium) \\
        \textbf{\sys} & \greencheck & \greencheck & \greencheck & \greencheck\tnote{1} &  \greencheck~(high) \\
        \bottomrule
    \end{tabular}}
    \begin{tablenotes}
    \item[1] The exact storage overhead depends on the scale parameter.
    \end{tablenotes}
\end{threeparttable}
    \label{tab:existing_approaches}
\end{table}

Except for \fump, all aforementioned \ac{FU} approaches rely on some form of gradient information to perform unlearning.
However, such approaches remain inefficient since gradients are expensive to store and recompute.
Approaches such as \su and \fump do not generalize to both client- and class-level unlearning.
\Cref{tab:existing_approaches} provides a comparison of existing \ac{FU} approaches and our work.
The popularity of gradient-based approaches for \ac{FU} motivates the following question:
\textit{what if one could succinctly compress all the gradient information such that it can be reused for fast gradient approximation for any downstream unlearning?}
This directly leads to our design of \sys, which, during \ac{FL} training, derives synthetic samples that accumulate gradient information in a compact form, enabling efficient unlearning.
\section{Design of \sys}
\label{sec:quickdrop_design}

\sys unlearns using \sga, but it does so on a synthetically generated dataset rather than with the original dataset.
This synthetic dataset is also used during recovery.
In \Cref{subsec:dd_for_unlearning}, we detail how the significantly smaller size of the synthetic data unlocks significant efficiency gains compared to standard \sga.
In \Cref{subsec:dd_formulation}, we formulate the process of generating synthetic data for unlearning and describe the algorithm for synthetic data generation, which operates in situ with FL training.
We discuss in \Cref{subsec:finetuning_distilled_data} how the synthetic data can be fine-tuned to boost model accuracy during recovery.
Finally, we summarize the end-to-end workflow of \sys in \Cref{subsec:quickdrop_final_summary}.

\begin{algorithm}[t!]
	\small
	\KwIn{Model parameters $\btheta$ trained via FL on training datasets $\{D_i\}_{i = 1}^N$, forget set $D_f$, local update steps $T$ and learning rate $\eta_{\btheta}$.}

    \textbf{Server executes:}
    
    \Indp
    Receive unlearning request for $D_f$

    Execute unlearning rounds using \fedavg($\md_f$, $\btheta$, \texttt{unlearn})

    Execute recovery rounds using \fedavg($\md \backslash \md_f$, $\btheta$, \texttt{recover})

    \Indm

    \textbf{\fedavg($\mz$, $\btheta$, \texttt{phase}):}

    \Indp
    
    Let $\{Z_i\}_{i=1}^N \leftarrow$ counterparts of $\mz$ with respective clients\footnotemark[1]

    $\btheta_{0,0} \leftarrow \btheta$
    
    \For{$k = 0,1, \cdots$ until convergence} 
	{
		
		\For{each client $i = 1, \cdots, N$ in parallel\footnotemark[2]}
		{
			Initialize $\btheta^{i}_{k,0} \gets \btheta_{k,0}$
			
			\For{$t = 0, \cdots, T-1$}
			{
					
					Sample batch $B\sim\mz_{i}$ and compute $\nabla\ml^{\mz_{i}}(\btheta_{k,t}^{i})$
          
					

                    \If{\texttt{phase} is \texttt{unlearn}}{
                    $\btheta_{k, t+1}^i \leftarrow \btheta_{k, t}^i + \eta_{\btheta} \nabla \ml^{\mz_i}(\btheta_{k, t}^i)$ 
                    }
                    \ElseIf{\texttt{phase} is \texttt{recover}}{
                    $\btheta_{k, t+1}^i \leftarrow \btheta_{k, t}^i - \eta_{\btheta} \nabla \ml^{\mz_i}(\btheta_{k, t}^i)$ 
                    }
					
				}
			}
			$\btheta_{k+1, 0} \leftarrow \sum_{i=1}^N \frac{|\mz_{i}|}{|\mz|} \btheta^i_{k,T}$
		}
		\caption{Unlearning using the \sga algorithm~\cite{wu2022federated}.}
		\label{algo:sga}
	\end{algorithm}
\footnotetext[1]{Only clients with non-empty $Z_i$ are required to participate.}
\footnotetext[2]{Clients can also be sub-sampled.}

\subsection{Synthetic Data Generation for Efficient Unlearning}
\label{subsec:dd_for_unlearning}
As detailed in \Cref{algo:sga}, \sga executes unlearning rounds on $\md_f$ followed by recovery rounds on $\md \backslash \md_f$.   
To significantly reduce the volume of data involved when executing an unlearning request, we generate synthetic samples that condense critical gradient information from the original training into a small synthetic dataset.
This process is also known as \ac{DD}~\cite{wang2018dataset}.
In our \ac{FL} setting, each client $i \in N$ independently synthesizes a synthetic dataset $S_i$ from its local dataset $D_i$ such that $|S_i| \ll |D_i|$. 
More specifically, each client locally generates a synthetic per-class counterpart $S_i^c$ of its original per-class data $D_i^c$.
Per-class generation of $ S_i $ enables \sys to perform both class- and client-level unlearning.
Similar to the training samples used by \ac{FL}, generated synthetic samples \textit{never leave the device}.
The unlearning algorithm $\mathcal{U}$ can thus be modified as $\btheta_f = \mathcal{U}(\btheta, S_f)$, where $S_f$ is the synthetic counterpart of the \forgetset $D_f$.
In other words, once the synthetic data is generated, \sys can serve any unlearning requests by executing unlearning rounds on $S_f$ and recovery rounds on $S \backslash S_f$.
To exemplify, when perform class-level unlearning for class $c$, we have $S_f := \cup_i S_i^c $ where client $i$ has class $c$.
Similarly, when performing client-level unlearning of the $i$-th client, we have $S_f := S_i$.
Since the synthetic data is orders of magnitude smaller in volume, the unlearning task can executed very efficiently as only a few unlearning and recovery rounds are required to unlearn and recover the knowledge, respectively.

\subsection{Formulating the Distillation Task}
\label{subsec:dd_formulation}

The goals of standard \ac{DD} differ from our task of synthetic data generation for unlearning.
We first formally describe standard \ac{DD} before formulating the task in the context of unlearning. 
Suppose we are given a training dataset $ \md $ from a distribution $P_\md$ containing $m_\md$ pairs of training images and class labels $\md=\{(\bx_i,y_i)\}|_{i=1}^{m_\md}$ where $\bx_i\in\mx\subset \mathbb{R}^d$, $y_i\in\{0,\dots,C-1\}$ and $C$ is the number of classes. 
Let $\phi$ be a learnable function (\eg a deep neural network) with parameters $\btheta$ that correctly predicts labels of images.
One can learn the parameters of this function by minimizing an empirical loss over the training set:
\begin{equation}
	\btheta^\md=\argmin_{\btheta}\ml^{\md}({\btheta})
	\label{eq.lossT}
\end{equation} where $\ml^{\md}({\btheta})=\frac{1}{m_\md}\sum_{(\bx,y)\in\md}\ell(\phi_{\btheta}(\bx),y)\;$, $\ell(\cdot,\cdot)$ is a task-specific loss (\ie cross-entropy) and $\btheta^\md$ is the minimizer of $\ml^{\md}$.
The generalization performance of the obtained model $\phi_{\btheta^\md}$ can be written as $\mathbb{E}_{\bx, y\sim P_{\mathcal{D}}}[\ell(\phi_{{\btheta}^\md}(\bx),y)]$.
\ac{DD} seeks to generate a small set of condensed synthetic samples with their labels, $\ms=\{(\bs_i,y_i)\}|_{i=1}^{m_\ms}$ where $\bs_i\in\mathbb{R}^d$ and $y_i\in\mathcal{Y}$ such that $m_\ms \ll m_\md$.
Similar to Eq. (\ref{eq.lossT}), one can train $\phi$ with these synthetic samples as follows:
\begin{equation}
	\btheta^\ms=\argmin_{\btheta}\ml^{\mathcal{S}}({\btheta})
	\label{eq.lossS}
\end{equation} where $\ml^\ms({\btheta})=\frac{1}{m_\ms}\sum_{(\bs,y)\in\ms}\ell(\phi_{\btheta}(\bs),y)$ and $\btheta^\ms$ is the minimizer of $\ml^\ms$.
The goal of standard \ac{DD} is to obtain $\ms$ such that the generalization performance of $\phi_{\btheta^\ms}$ is as close as possible to $\phi_{\btheta^\md}$, \ie, 
\begin{equation}
\mathbb{E}_{\bx,y \sim P_{\mathcal{D}}}[\ell(\phi_{\btheta^\md}(\bx),y)]\simeq\mathbb{E}_{\bx,y \sim P_{\mathcal{D}}}[\ell(\phi_{\btheta^\ms}(\bx),y)]
\end{equation}
over the real data distribution $P_{\mathcal{D}}$.

\subsubsection{Formulating \ac{DD} for \ac{FU}.}
While the above \ac{DD} formulation may work for \ac{FU}, generating general-purpose synthetic samples is a compute intensive process, requiring many optimization iterations~\cite{zhang2023accelerating}. 
To significantly reduce the computational overhead for clients, we reformulate the synthetic data generation task to specifically target unlearning instead of general-purpose \ac{DD}.
We refer to the outcome of $ \mathcal{U}(\btheta, \md_f)$ by $\btheta_{\md_f}$ and denote the outcome of $ \mathcal{U}(\btheta, \ms_f)$ by $\btheta_{S_f}$.
The goal of \ac{DD} in \sys is to generate synthetic data such that the generalization performance of the unlearned model $\phi_{\btheta_{\ms_f}}$ is as close as possible to unlearned model $\phi_{\btheta_{\md_f}}$, \ie, 
\begin{equation}
\mathbb{E}_{\bx,y \sim P_{\mathcal{D}}}[\ell(\phi_{\btheta_{\ms_f}}(\bx),y)]\simeq\mathbb{E}_{\bx,y \sim P_{\mathcal{D}}}[\ell(\phi_{\btheta_{\md_f}}(\bx),y)]
\end{equation}
over the global data distribution $P_{\mathcal{D}}$.

\subsubsection{\ac{DD} with Gradient Matching.}
\label{subsubsec:dd_with_gradient_matching}
We aim to generate synthetic samples such that the impact of forgetting the target synthetic data is similar to the effect of forgetting the target original data.
When using \sga as the unlearning algorithm $\mathcal{U}$, intuitively, the synthetic data should be such that the gradients are as close as possible to the original data, thereby following the reverse trajectory along the optimization path when ascending.
Additionally, generating such synthetic data must be local to the client, avoiding any transfer of private local information.
Inspired by the dataset condensation algorithm of Zhao \etal~\cite{zhao2021dataset} in centralized settings, we achieve this by matching gradients obtained over training and synthetic data over several time steps during the execution of \ac{FL} training, locally at each client.
Precisely, each client obtains its synthetic data $\ms_i$ by minimizing the following:
\begin{equation}
	\min_{\ms_i} \sum_{k=0}^{K-1} \sum_{t=0}^{T-1} d(\nabla_{\btheta}\ml^{\ms_i}(\btheta_{k,t}^i),\nabla_{\btheta}\ml^{\md_i}(\btheta_{k,t}^i))
	\label{eq.optgrad}
\end{equation}
where $d(.;.)$ is a function that measures the distance between the gradients for $\ml^{\ms_i}$ and $\ml^{\md_i}$ w.r.t $\btheta$, $K$ denotes the total number of \ac{FL} global rounds and $T$ represents the number of local steps within each round.
At each time step, the synthetic data samples are updated by running a few steps of an optimization algorithm \texttt{opt-alg}, \eg, \ac{SGD}:
\begin{equation}
	\label{eq:synthetic_data_update}
	\ms_i \leftarrow \texttt{opt-alg}_{\ms_i}(d(\nabla_{\btheta}\ml^{\ms_i}(\btheta_{k,t}^i),\nabla_{\btheta}\ml^{\md_i}(\btheta_{k,t}^i)),\varsigma_{\ms},\eta_{\ms})
\end{equation}
where $\varsigma_{\ms}$ and $\eta_{\ms}$ are the number of steps and the learning rate.
In other words, the synthetic data absorbs the gradient information along the optimization trajectory of the \ac{FL} model $\btheta_{k,t}^i$.
The synthetic data, therefore, can be interpreted as a \textit{compact compression} of all the gradient information, which is reused for fast gradient approximation in downstream unlearning. 

We detail this procedure to generate synthetic data in \Cref{algo:quickdrop_dd}.
Each client initializes its per-class synthetic dataset $\ms_i^c$ by randomly picking samples from the original dataset $D_i^c$ (lines 2-7).
In \ac{FL} settings, clients typically have a highly unbalanced number of samples per class (non-IIDness)~\cite{zhu2021federated}.
Thus, we set the number of synthetic samples per class proportionately to the original per-class dataset size through the scale parameter $s$ as $|\ms_i^c| = \lceil |\md_i^c| / s \rceil$.
The \textsc{ceil} function ($\lceil . \rceil$) ensures at least one sample per class in synthetic datasets if the class exists within $\md_i$.
Increasing $ s $ decreases the number of synthetic samples but also reduces the accuracy after recovery since samples contain less information.
We set the scale parameter to \num{100}, resulting in a synthetic dataset which is 1\% in volume of the original dataset.
We found this value of $ s $ to yield a reasonable balance between efficiency and effectiveness.
While this incurs some storage overhead for clients, it is marginal compared to the volume of the original dataset.
We also experiment with different values of $ s $ in~\Cref{subsec:exp_scale_factor}.

Synthetic data generation happens in situ with \ac{FL} training, resulting in a negligible compute overhead for clients (also see~\cref{subsec:exp_dd_finetuning}).
During each local update step, clients sample mini-batch pairs $B^{\md_i}$ and $B^{\ms_i}$ from their original and synthetic datasets, respectively (line 12).
They evaluate respective gradients on the current model parameter $\btheta_{k, t}^i$ (lines 13 and 14).
While the clients use the gradient on the original data ($\nabla\ml^{\md_i}$) to perform the \ac{FL} local update (line 17), the gradient is also matched with the gradient on the synthetic data ($\nabla\ml^{\ms_i}$) (line 15).
The synthetic samples corresponding to the mini-batch $B^{\ms_i}$ are then updated through the execution of \texttt{opt-alg} for $\varsigma_{\ms}$ steps.
The server executes parameter aggregation like standard \ac{FL} (line 18).
By the end of \ac{FL} training, \sys generates per-client synthetic datasets $\{\ms_i\}_{i = 1}^N$ along with the trained \ac{FL} global model $\btheta_{K,0}$.
Finally, we remark that \sys employs a class-wise gradient matching, similar to \cite{zhao2021dataset}, wherein the update of synthetic samples in $B^{\ms_i}$ takes place class-wise.
We omitted it from \Cref{algo:quickdrop_dd} for presentation clarity.


\begin{algorithm}[t!]
	\small
	\KwIn{Training datasets $\{D_i\}_{i = 1}^N$, FL global rounds $K$, local update steps $T$, number of local update steps $\varsigma_{\ms}$ and learning rate $\eta_{\ms}$ for synthetic samples, model learning rate $\eta_{\btheta}$ and scale parameter $s$.}
	
	$\vartriangleright$ Initialization
	
	\For{each client $i = 1, \cdots, N$}
	{
		\If{client $i$ has class $c$}
		{
			$\ms_i^c \leftarrow$ Randomly pick $\lceil |\md_i^c|/s \rceil$ samples from $\md_i^c$
		}
		\Else
		{
			$\ms_i^c \leftarrow \emptyset $
		}
		
		$\ms_i \leftarrow \cup_c S_i^c$
	}
	
	\For{$k = 0, \cdots, K-1$} 
	{
		
		\For{each client $i = 1, \cdots, N$ in parallel}
		{
			Initialize $\btheta^{i}_{k,0} \gets \btheta_{k,0}$
			
			\For{$t = 0, \cdots, T-1$}
			{
					
					Sample minibatch pairs $B^{\md_i}\sim\md_i$ and $B^{\ms_i}\sim\ms_i$
					
					Compute $\nabla\ml^{\md_i}(\btheta_{k,t}^{i})$ on  $B^{\md_i}$
					
					Compute $\nabla\ml^{\ms_i}(\btheta_{k,t}^{i})$ on $B^{\ms_i}$
					
					Update $\ms_i$ using $d(\nabla\ml^{\md_i}$,$\nabla\ml^{\ms_i})$ \hfill $\rhd$ \cref{eq:synthetic_data_update}
					
					$\vartriangleright$ Update \ac{FL} model using the above gradient
					
					$\btheta_{k, t+1}^i \leftarrow \btheta_{k, t}^i -\eta_{\btheta} \nabla \ml^{\md_i}(\btheta_{k, t}^i)$
					
				}
			}
			$\btheta_{k+1, 0} \leftarrow \sum_{i=1}^N \frac{|\md_i|}{|\md|} \btheta^i_{k,T}$
		}
		\KwOut{$\{\ms_i\}_{i=1}^N, \btheta_{K,0}$}			
		\caption{Generating synthetic data during \ac{FL} training in \sys.}
		\label{algo:quickdrop_dd}
	\end{algorithm}

\subsection{Enhancing Recovery for \sga}
\label{subsec:finetuning_distilled_data}
Once the synthetic data is generated on each client, \sys employs \sga to unlearn the \forgetset $D_f$ through $S_f$. 
However, the model must still undergo the recovery phase on the \retainset $D \backslash D_f$ to restore the performance on the remaining data.
Recovery is akin to original \ac{FL} training and hence requires high-quality samples.
While recovery in \sys can be conducted using $D \backslash D_f$, the volume of remaining data can be quite large, thus significantly affecting the efficiency.
On the other hand, while we can use $S \backslash S_f$ to achieve efficient recovery, the restoration quality is not the same as when using $D \backslash D_f$. 
This is because the synthetic data generation was not targeted towards generalization.
We address this issue through the following two amendments.

\subsubsection{Data augmentation with original samples}
\label{subsubsec:data_augmentation}
We found that even including a few original samples in the synthetic datasets during recovery can significantly boost the restoration performance.
In our experimental setting, we randomly select samples from the original dataset with the same size as the synthetic dataset (\ie, synthetic : selected = 1:1). 
Since the size of the synthetic dataset is $1\%$ compared to the original dataset, the relative size of the mixed dataset is only $2\%$ after combining selected samples.
Therefore, its influence on storage overhead is still marginal.

\subsubsection{Fine-tuning (optional)}
\label{subsubsec:finetuing}
One can optionally enhance the recovery accuracy by conducting additional optimization steps on the previously generated synthetic data, referred to as fine-tuning.
We remark that even without fine-tuning, \sys already demonstrates excellent unlearning performance.
For fine-tuning, we use the distillation algorithm of Zhao \etal \cite{zhao2021dataset} which targets generalization.
More specifically, the algorithm performs gradient matching but across thousands of parameter initializations of the deep neural network $\phi$, necessary to achieve good generalization.  
In~\cref{subsec:exp_dd_finetuning}, we show that only \num{200} of such fine-tuning steps are sufficient to make the recovery accuracy nearly match that of the oracle (\retrainor).
Fine-tuning is independently executed by each client on their synthetic dataset $\ms_i$.

\subsection{End-to-end Workflow of \sys}
\label{subsec:quickdrop_final_summary}
Finally, we summarize the end-to-end workflow of \sys, also depicted in Figure~\ref{fig:quickdrop}.
Our algorithm involves the following five steps:
\begin{enumerate}
	\item \textbf{FL training + synthetic data generation --}  Clients initially train a global model via standard \ac{FL} while also deriving synthetic data samples for unlearning. This procedure is described in \Cref{algo:quickdrop_dd}.
	\item \textbf{Data augmentation + fine-tuning --} Clients augment their synthetic sets with a very small number of original data samples and optionally conduct some fine-tuning steps to improve the quality of synthetic data for recovery.
	\item \textbf{Unlearning --} When an unlearning request arrives at the server, clients engage in \sga-based unlearning using their synthetic datasets. 
	If the request concerns class-level unlearning, all clients containing the specific target class $c$ are involved, \ie $S_f := \cup_i S_i^c$ where client $i$ has class $c$. 
	If the request concerns client-level unlearning, only the specific target client is involved, \ie $S_f := S_i$.
	\item \textbf{Recovery --} All clients with remaining data after removing the \forgetset engage in the recovery phase. 
	These clients perform standard \ac{FL}-like training (\ie using \sgd) on their synthetic datasets.
    \item \textbf{Relearning --} Clients are able to efficiently relearn previously unlearned knowledge by performing standard \ac{FL}-like training (\ie using \sgd) on $S_f$.
\end{enumerate}

\section{Evaluation}
\label{sec:experiments}

We conduct extensive evaluations to assess the computational efficiency and performance of \sys using three standard datasets and against five \ac{FU} baselines.
Our evaluations cover the performance of a single unlearning request (\Cref{subsec:single_req}), multiple unlearning requests (\Cref{subsec:exp_sequential_unlearning}), the effect of fine-tuning and the scale factor (\Cref{subsec:sensitivity_analysis}), the performance in larger networks (\Cref{subsec:exp_scalability}), client-level unlearning (\Cref{subsec:client_unlearn}), the performance of relearning (\Cref{subsec:relearning}) and the computational overhead incurred by \sys (\Cref{subsec:exp_compute_overhead}).




\subsection{Experimental Setup}
\label{subsec:experiment_setups}

We evaluate the performance of \sys on three standard image classification datasets: \mnist~\cite{lecun1998mnist}, \cifar~\cite{krizhevsky2009learning}, and \svhn~\cite{netzer2011reading}.
For all datasets, we use a ConvNet as the deep neural network backbone~\cite{gidaris2018dynamic}. 
Its modular architecture contains $D$ duplicate blocks, and each block has a convolutional layer with $W$ ($3 \times 3$) filters, a normalization layer $N$, an activation layer $A$, and a pooling layer $P$, denoted as $[W, N, A, P]\times D$. The default ConvNet (unless specified otherwise) includes $3$ blocks, each with \num{128} filters, followed by InstanceNorm, ReLU, and AvgPooling modules.
A linear classifier follows the final block.

All experiments are conducted on a machine equipped with an i5-10600K CPU and an RTX 2060 GPU.
All source code is made available in a GitHub repository. \footnote{See \url{https://github.com/sacs-epfl/quickdrop}.}

\textbf{Federated Learning.}
To generate local datasets with varying data heterogeneity, we adopt the Dirichlet distribution-based approach from previous works~\cite{hsu2019measuring,zhu2021data}.
The degree of non-IIDness is controlled by a parameter $\alpha \in [0, \infty)$, with lower values corresponding to higher heterogeneity.
We fix $ \alpha = 0.1 $ for all our experiments involving non-IID data.
This value is commonly used in literature and results in highly non-IID data partitions across clients.
We also conduct experiments with IID distributions in~\Cref{subsec:client_unlearn}.
We conduct experiments with \num{10} and \num{20} total clients, except in Section~\ref{subsec:exp_scalability}, where we assess the performance of \sys in a network with $100$ clients.
All clients participate in each round of FL training and unlearning unless mentioned otherwise.

\textbf{Hyperparameters.} In all experiments, we run $K=200$ global rounds for \ac{FL} training with $50$ local steps per round \ie $T = 50$.
We use a mini-batch size of $256$ for \ac{FL} training \ie to generate gradients on real images.
We use \ac{SGD} as the client local optimizer and a learning rate $\eta_{\btheta} = 0.01$ for \ac{FL} training.
The learning rates are separately tuned for unlearning and recovery, which are $0.02$ and $0.01$, respectively.
For all experiments with \sys, we perform a single round of unlearning and two rounds for recovery, which we found to be the minimum for the model to sufficiently unlearn particular knowledge and restore the performance on the remaining data. 
All experiments use data augmentation for recovery (\Cref{subsubsec:data_augmentation}), but we disable fine-tuning for all experiments (\ie, $ F = 0 $), except for the experiments reported in Section~\ref{subsec:exp_dd_finetuning}.

\textbf{Synthetic data generation.}
For synthetic data generation, we set $\varsigma_{\ms}=1$, $\eta_{\ms} = 0.1$ and use \sgd as the \verb|opt-alg| algorithm following ~\cite{zhao2021dataset}.
When fine-tuning using the algorithm of Zhao \etal~\cite{zhao2021dataset}, we use their default hyperparameters while varying the number of fine-tuning steps $ F $.
More specifically, we vary the number of outer-loop steps while keeping the inner-loop steps fixed at $50$.
We initialize the synthetic samples $\{\ms_i\}_{i=1}^N$ as randomly selected real training samples from the original local client dataset.
We found this to be more effective in our setting than initializing these samples from Gaussian noise.
We use the same distance function $d$ for gradient matching as in~\cite{zhao2021dataset} and, except for the experiments in~\Cref{subsec:exp_scale_factor}, fix the scale parameter $s=100$ for all experiments.

\textbf{Baselines.}
We compare the performance of \sys with the five \ac{FU} baselines discussed in \Cref{subsec:existing_algos}.
These are \retrainor (the retraining oracle), \sgaor (\sga on the original dataset)~\cite{wu2022federated},  \federaser~\cite{liu2021federaser}, \fump~\cite{wang2022federated} and \su~\cite{gao2022verifi}.
Our baselines thus cover a diverse array of \ac{SOTA} techniques on both client- and class-level unlearning.
We focus on CV tasks to ensure compatibility with our selected baselines.
Many of our baselines primarily demonstrate their performance on CV tasks, and some, like \fump, are specifically designed for CNNs due to their use of channel pruning techniques.

\textbf{Metrics.}
We report testing accuracy as the Top-1 accuracy achieved on the specific testing data of each dataset (\eg class-wise testing accuracy when doing class-level unlearning).
For a more fine-grained comparison, we also report the accuracy on the \forgetset, referred to in this section as the \emph{\fset}, and on the \retainset, referred to as the \emph{R-Set}.
The goal of any \ac{FU} approach is to match the accuracy of \retrainor (\ie oracle's performance) on both the \fset and \rset.
We run each experiment five times and report averaged values.

\begin{figure}[t]
    \centering
    \includegraphics[width=.95\linewidth]{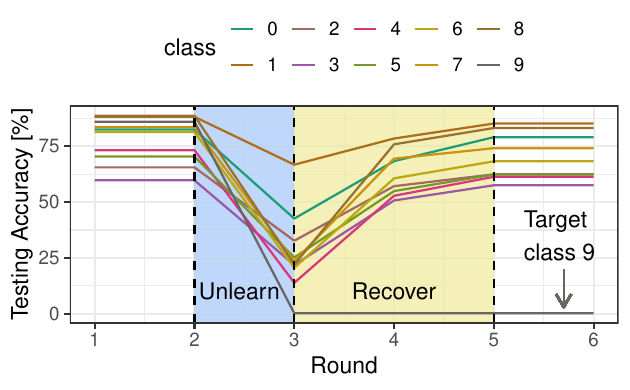}
    \caption{Class-wise testing accuracy on the \cifar dataset when unlearning class $9$. \sys successfully unlearns class $9$ while retaining good performance on the remaining classes after recovery.}
    \label{Fig:Accuracy_one_request}
\end{figure}

\begin{table*}[t]
\centering
\caption{The accuracy and computation cost of \sys and \ac{FU} baselines under class-level unlearning on the \cifar dataset, with non-IID data distributions ($\alpha=0.1$) and in a $10$-client network.
	The F-Set and R-Set denote the accuracy on the \forgetset and the \retainset.
	The speedup is measured with respect to \retrainor.
	}
\scalebox{0.96}{
\begin{tabular}{c|cc|ccc|cc|ccc|cc}
\toprule
\textbf{Stage} & \multicolumn{5}{c}{Unlearning} & \multicolumn{5}{|c}{Recovery} & \multicolumn{2}{|c}{Unlearn. + Recovery} \\
\cmidrule(r){2-6} \cmidrule(r){7-11}  \cmidrule(r){12-13}
\multicolumn{1}{c}{} & \multicolumn{2}{|c}{Accuracy}   & \multicolumn{3}{c}{Computation Cost} & \multicolumn{2}{|c}{Accuracy} & \multicolumn{3}{c}{Computation Cost} & \multicolumn{2}{|c}{Total} \\
\cmidrule(r){2-3} \cmidrule(r){4-6} \cmidrule(r){7-8} \cmidrule(r){9-11}  \cmidrule(r){12-13}
\textbf{\ac{FU} approach} & \fset & \multicolumn{1}{c|}{R-Set} & Round & Time (s) & Data Size & \fset & \multicolumn{1}{c|}{R-Set} & Round & Time (s) & Data Size & Time (s) & Speedup \\
\cmidrule(r){1-1} \cmidrule(r){2-3} \cmidrule(r){4-6} \cmidrule(r){7-8} \cmidrule(r){9-11} \cmidrule(r){12-13}
\retrainor & $0.81\%$ & $74.95\%$ & $30$ & $7239.58$ & \num{45000} & --- & --- & --- & --- & --- & $7239.58$ & $1 \times$ \\
\federaser & $0.02\%$ & $22.01\%$ & $10$ & $2637.42$ & \num{45000} & $0.01\%$ & $69.67\%$ & $3$ & $764.83$ & \num{45000} & $3402.25$ & $2.12 \times$ \\
\sgaor & $0.75\%$ & $48.69\%$ & $2$ & $495.17$ & \num{5000} & $1.03\%$ & $74.83\%$ & $2$ & $551.33$ & \num{45000} & $1046.50$ & $6.92 \times$ \\
\fump & $0.12\%$ & $11.58\%$ & $1$ & $61.36$ & \num{50000} & $0.09\%$ & $73.96\%$ & $4$ & $953.62$ & \num{45000} & $1014.98$ & $7.13 \times$ \\
\sys & $0.82\%$ & $37.68\%$ & $1$  & $5.03$ & \num{100} & $0.85\%$ & $70.48\%$ & $2$ & $10.58$ & \num{900} & $\textbf{15.61}$ & $\textbf{463.7} \times$ \\
\bottomrule
\end{tabular}}
\label{Tab:combined_accuracy_computation_comparison}
\end{table*}

\subsection{Performance of a Single Unlearning Request}
\label{subsec:single_req}


In this section, we will show the effectiveness of \sys by evaluating the efficiency and effectiveness when doing a single unlearning request.



\subsubsection{Unlearning a Single Class}
We first quantify the change in testing accuracy of target and non-target classes
after the unlearning and recovery stages, using the \cifar dataset and $10$ clients.
The network collaboratively unlearns from the model the knowledge corresponding to class $9$ (digit \enquote{$9$}) by performing one round of unlearning and two rounds of recovery.
Figure~\ref{Fig:Accuracy_one_request} shows the testing accuracy for each class during six rounds in different colors when unlearning with \sys.
When \sys starts the unlearning stage (round \num{2}), we observe a rapid accuracy drop on the target class while the accuracy of non-target classes decreases as well.
This is because \ac{SGA} introduces some noise that affects non-target classes, even though the model parameters changed by \ac{SGA} are mainly for unlearning the knowledge of the target class.
We observe that even \emph{a single unlearning round is sufficient to unlearn all knowledge of the target class} (its accuracy drops to $0.82\%$), and executing additional unlearning rounds is futile.
Therefore, we execute a single unlearning round for \sys in all remaining experiments.
The recovery stage starts immediately after the unlearning stage (round \num{3}).
Figure~\ref{Fig:Accuracy_one_request} shows that the accuracies of non-target classes after two recovery rounds are almost restored to their original values.
We observed that executing additional recovery rounds did not improve the performance of non-target classes. 
In the following section, we validate that the accuracies obtained by \sys are consistent with those from \retrainor, demonstrating effective unlearning.

\subsubsection{Accuracy against Baselines.}
Next, we compare the performance of \sys with our baselines on \cifar when unlearning a single class.
Table~\ref{Tab:combined_accuracy_computation_comparison} shows the accuracy on the \fset and R-Set for different \ac{FU} approaches and after each stage (unlearning and recovery).
We remark that there is no recovery stage for \retrainor.
Table~\ref{Tab:combined_accuracy_computation_comparison} shows that, after the unlearning stage, all approaches effectively eliminate the knowledge of the target class in the model as the \fset accuracy matches that of \retrainor which achieves $0.81\%$. 
\sys achieves $0.82\%$ on the F-set, demonstrating the effectiveness of synthetic samples for unlearning.
After recovery, all baselines restore the accuracy of the R-Set close to the values achieved with \retrainor.
The accuracy of \sys after recovery, $70.48\%$, is slightly lower than that of the baselines except for \federaser.
This is because the synthetic samples do not perfectly represent the original dataset, as they are optimized for unlearning.
However, additional fine-tuning of synthetic datasets can close this gap at the cost of additional computation (see Section~\ref{subsec:exp_dd_finetuning}).
Nonetheless, we conclude that \sys effectively unlearns data samples with minimal impact on the performance of remaining samples.
More importantly, \sys unlearns significantly more efficiently than the baselines, thanks to the synthetic samples, as we show in \Cref{subsubsec:comp_eff}.

\begin{figure}[b]
    \centering
    \includegraphics[width=\linewidth]{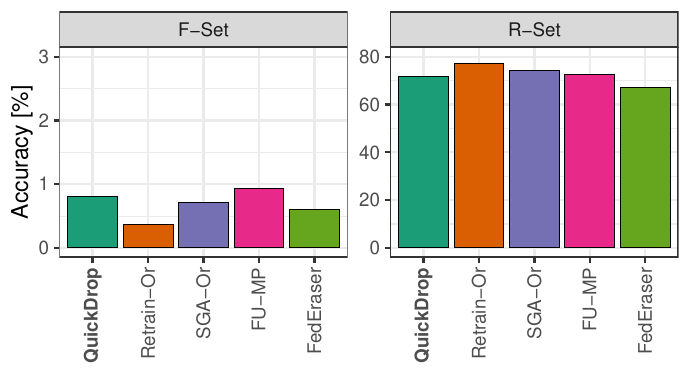}
	\caption{The membership inference attack (MIA) accuracy of all baselines after unlearning on the \cifar dataset with $10$ clients and non-IID partitioning.
    }
    \label{fig:mia_accuracy}
\end{figure}

\begin{figure*}[t]
    \centering
    \includegraphics[width=.9\linewidth]{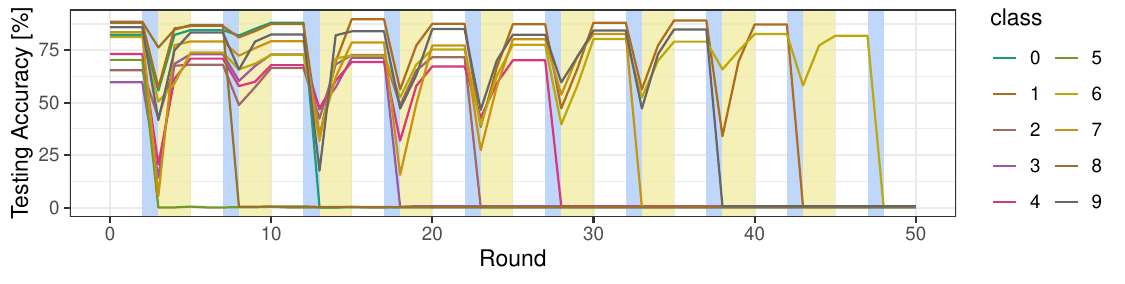}
    \caption{The accuracy of each class with sequential unlearning requests on \cifar and with $ \alpha=0.1 $. We unlearn a random class every five rounds and highlight the unlearning and recovery stages in blue and yellow shades, respectively. The class unlearning order is $[5, 8, 0, 3, 2, 4, 7, 9, 1, 6]$.
    \label{fig:sequential_unlearn}}
\end{figure*}

\subsubsection{Membership Inference Attack.}
To further assess the effectiveness of unlearning with \sys, we conduct a \ac{MIA} on the unlearned model, which follows related work on \ac{MU}~\cite{chen2021machine}.
The \ac{MIA} aims to determine whether a particular sample is included in the model's knowledge.
We implement the \ac{MIA} according to the settings in \cite{golatkar2021mixed} and measure how often the attack model classifies a data sample from deleted data as a training sample of the unlearned model, serving as an alternative metric to test accuracy.
\Cref{fig:mia_accuracy} shows these \ac{MIA} accuracies on the F-Set and R-Set after the unlearning algorithm terminates.
The performance of \retrainor can be considered optimal since the trained model has never seen the unlearned samples.
For all approaches, the \ac{MIA} accuracy on the \fset is below 1\%.
We also observe that the \ac{MIA} accuracy of \sys on the R-Set ($71.62\%$) is competitive with that of the baselines ($67.28-74.21\%$) while the oracle achieves $77.25\%$.
Thus, \sys effectively unlearns knowledge from a trained model, evidenced by its \ac{MIA} performance.

\subsubsection{Computation Efficiency.}
\label{subsubsec:comp_eff}
Table~\ref{Tab:combined_accuracy_computation_comparison} also shows the computational cost for unlearning and recovery in terms of rounds, time required, and the number of data samples involved in executing these rounds.
The computation costs (round and time) correspond to the attainment of convergence in the specific phase (unlearning or recovery) for each baseline.
We observe significant differences in computation cost between the evaluated \ac{FU} approaches.
Since \ac{DD} reduces the number of samples for each client, the unlearning stage in \sys only takes \SI{5.03}{\second}, and \SI{10.58}{\second} in the recovery stage; both stages are completed in just \SI{15.61}{\second}.
This efficiency is because a round of unlearning and recovery with \sys only involves \num{100} and \num{900} data samples, respectively.
Although \sgaor only needs two rounds each to unlearn and recover, it takes much longer (\SI{1046.50}{\second}) in total than \sys (\SI{15.61}{\second}) since it must operate on the complete original data (\num{5000} data samples in the unlearning stage and \num{45000} samples in the recovery stage).
While \retrainor is the simplest \ac{FU} approach with the highest R-Set accuracy after unlearning, its computational time renders this approach infeasible in many scenarios, which is \num{6.92}$\times$ higher than \sgaor and \num{463.7}$\times$ higher than \sys. 
We note that \fump employs a different technique than other baselines, \ie model pruning, while its recovery stage is the same.
Since model pruning only depends on information obtained from inference, which can be done relatively quickly, the unlearning is relatively fast (\SI{61.36}{\second}).
However, the recovery is still slow, resulting in a total time of \SI{1014.98}{\second} in comparison to \SI{15.61}{\second} for \sys.
Thus, \sys remains $65 \times$ faster than \fump.

Although the computation efficiency of \federaser improves over \retrainor, it stays relatively low (\SI{3402.25}{\second}) due to the model update calibration step, which requires extra model training on all the remaining data of non-target classes.
Consequently, \sys also is significantly faster than \federaser ($218\times$).
We note that \federaser also incurs significant storage costs to retain historical model updates, which increases linearly with the number of clients and executed FL rounds.
Thus, from~\Cref{Tab:combined_accuracy_computation_comparison}, we conclude that \sys achieves quick unlearning and recovery, boasting a speedup of $463.7\times$ over \retrainor and $65-218\times$ compared to other baselines.

\subsection{Performance of Multiple Unlearning Requests}
\label{subsec:exp_sequential_unlearning}

In real-world settings, clients may continually launch unlearning requests.
Therefore, we go beyond existing work on \ac{FU} and evaluate the performance of \sys with sequential class unlearning requests.
Figure~\ref{fig:sequential_unlearn} shows the accuracies when sequentially unlearning all ten \cifar classes in random order.
We observe that the unlearning phase for each target class results in low testing accuracy as before.
Although the accuracies of non-target classes also drop after the unlearning stage, they are rapidly restored in the recovery stages while leaving the \textit{low accuracy of the unlearned classes unaffected}.
Therefore, Figure~\ref{fig:sequential_unlearn} shows the capability of \sys in executing multiple unlearning requests.

\subsection{Sensitivity Analysis}
\label{subsec:sensitivity_analysis}
We now experiment with the most important parameters of \sys and analyze their sensitivity on the achieved accuracy and efficiency.
We first explore the impact of doing additional fine-tuning steps and then analyze the impact of the scale factor.

\begin{figure}[t]
	\centering
	
	\includegraphics[width=\linewidth]{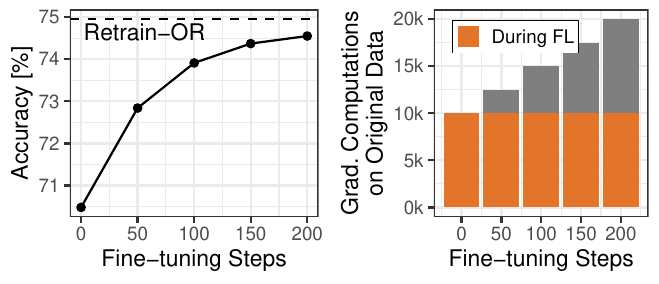}
	\caption{The accuracy on the R-Set after recovery (left) and the number of gradient computations performed on original data (right) when doing additional fine-tuning steps on \cifar.
		The portion in orange corresponds to \ac{FL} training, while the gray portion corresponds to new gradients computed for fine-tuning.
		\sys nearly matches the accuracy of \retrainor at an extra gradient cost no higher than that of \ac{FL} training.
	}
	\label{Fig:sensitivity_FT}
\end{figure}

\subsubsection{Impact of Fine-tuning}
\label{subsec:exp_dd_finetuning}
As discussed in~\Cref{subsec:dd_for_unlearning}, while the synthetic samples are optimized for unlearning, they might not perform sufficiently well during recovery.
To offset this, \sys allows clients to perform additional fine-tuning steps ($F$) to refine all synthetic samples using the algorithm of Zhao \etal~\cite{zhao2021dataset}.
Figure~\ref{Fig:sensitivity_FT} (left) shows the accuracy of \sys on the R-Set after the recovery stage when doing more fine-tuning (\ie, increasing $ F $ from \num{0} to \num{200}).
We also show with a dashed horizontal line the accuracy of \retrainor (74.95\%), which we consider optimal.
We observe an increase in accuracy as $ F $ increases: from 70.48\% at $ F = 0 $ to 74.55\% at $ F = 200 $.
More fine-tuning, however, comes at the cost of additional computation.
Figure~\ref{Fig:sensitivity_FT} (right) shows the total number of gradient computations performed on the original CIFAR-10 dataset for one client.
This figure marks the portion of gradients generated during \ac{FL} training in orange, while the portion of gradients in gray corresponds to the ones computed for fine-tuning.
As $F$ increases to \num{200}, the number of gradients computed for fine-tuning ($10$k) match that of \ac{FL} training ($10$k).
Consequently, with an extra gradient cost no higher than that of \ac{FL} training, \sys clients can achieve performance parity with \retrainor through fine-tuning.


\subsubsection{Impact of the Scale Parameter}
\label{subsec:exp_scale_factor}
\begin{figure}[t]
    \centering

    \includegraphics[width=\linewidth]{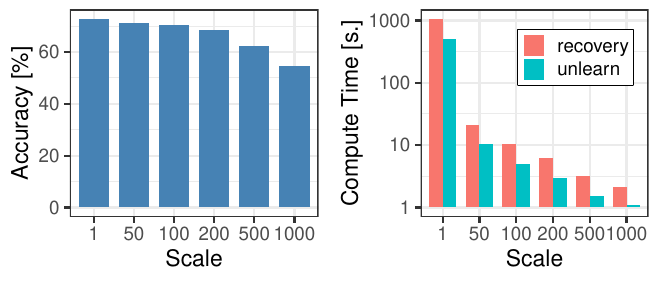}
    \caption{The accuracy of the R-Set after recovery (left) and the total compute time required for unlearning and recovery, for different scales $ s $.}
    \label{Fig:sensitivity_scale}
\end{figure}

The scale parameter $ s $ determines the ratio of original to synthetic samples per class and has a key impact on the computational efficiency and accuracy of \sys.
We explore the impact of this parameter by varying $ s $ from \num{1} to \num{1000}, using the same setup as the experiment described in~\Cref{subsec:single_req} (with the \cifar dataset and 10 clients).
\Cref{Fig:sensitivity_scale} (left) shows the accuracy on the R-Set after recovery, for the considered values of $ s $.
As $ s $ increases, the accuracy on the R-set after recovery decreases.
This is because increasing $ s $ results in lower number of synthetic samples and consequently more compression, inducing difficulty in unlearning.
This decrease in accuracy is less noticeable when ranging $ s $ between \num{1} and \num{200}.
With $ s = 1$ and $ s = 100 $, we observe an accuracy of 72.67\% and 70.48\%, respectively.
The accuracy achieved when using only original samples is 74.83\% (corresponding to \sgaor).
However, for $s > 200$, accuracy drops rapidly, reaching just 54.69\% at $s = 1000$, where clients often have only one synthetic sample per class. 
Nonetheless, for this experiment setting, \sys is able to retain high accuracies when $ s \leq 100 $.

\Cref{Fig:sensitivity_scale} (right) shows the total computation time (in logarithmic scale) incurred by all clients to execute the unlearning and recovery stages for different values of $ s $.
As $ s $ increases, the time to execute these stages decreases significantly.
For instance, while the unlearning time for $s = 1$ is just over $8$ minutes, it drops to only 5 seconds for $s = 100$ and as low as $1$ second for $s = 1000$. 
Similarly, the recovery time decreases from $17.4$ minutes for $s = 1$ to $10.6$ seconds for $s = 100$.
This decrease is because with an increasing value of $ s $, clients will have less samples, thus decreasing the compute costs.
Based on these results, we use $s = 100$ in all our experiments which achieves a good trade-off between accuracy and compute efficiency.

\begin{table}[t]
	\centering
	\caption{The accuracy and computation cost of different \ac{FU} approaches with $100$ clients on the \svhn dataset for \textit{class-level} unlearning. 
		We report the performance when the algorithm terminates \ie after both unlearning and recovery.  
}
	
\scalebox{1}{
	\begin{tabular}{c|cc|cc}
		\toprule
		\multicolumn{1}{c}{\textbf{Metric}} & \multicolumn{2}{|c}{Accuracy}   & \multicolumn{2}{c}{Computation Cost} \\
		\cmidrule(r){2-3} \cmidrule(r){4-5} 
		\textbf{\ac{FU} approach} & \fset & \multicolumn{1}{c}{R-Set} & Time (s) & Speedup\\
		\cmidrule(r){1-1} \cmidrule(r){2-3} \cmidrule(r){4-5}
		\retrainor & $0.34\%$ & $88.39\%$ & $10483.51$ & $1\times$ \\
		\federaser & $0.38\%$ & $82.98\%$ & $2447.80$  & $4.28\times$ \\
		\sgaor & $0.66\%$ & $86.47\%$ & $1276.13$ & $8.21\times$\\
		\fump & $0.73\%$ & $85.63\%$ & $1927.43$ & $5.43\times$ \\
		\sys & $0.81\%$ & $84.96\%$ & $32.09$ & $\textbf{326.69}\times$\\
		\bottomrule
\end{tabular}}
\label{Tab:scalability_on_large_scale}
\end{table}

\subsection{Performance in Larger Networks}
\label{subsec:exp_scalability}

We next analyze the unlearning performance of \sys and other baselines for class-level unlearning in a \num{100}-client network with the \svhn dataset~\cite{netzer2011reading}.
Notably, our chosen network size surpasses the typical scale explored in related \ac{FU} research~\cite{liu2021federaser,wu2022federated}.
\svhn is a larger dataset compared to \mnist and \cifar, containing 10 classes and more than \num{600000} samples.
In each round, the server selects a random 10\% subset of clients to update the model during training and recovery, while the participation rate for unlearning is 100\%.
\Cref{Tab:scalability_on_large_scale} shows for each approach the accuracy on the \fset and R-Set when unlearning class $9$.
Even with $100$ clients, \sys effectively unlearns class knowledge, achieving F-set accuracy of $0.81\%$ while \retrainor achieves $0.34\%$.
Furthermore, \sys shows competitive accuracy on the R-Set ($84.96\%$) compared to the baselines ($82.98-86.47\%$), despite the large number of clients and samples in the training dataset.
While \retrainor achieves slightly higher R-Set accuracy ($88.39\%$), \sys shows a massive $326.69\times$ speedup over \retrainor.

\subsection{Client-level Unlearning}
\label{subsec:client_unlearn}
Our previous evaluations focused on class-level unlearning.
We now evaluate the effectiveness of \sys on client-level unlearning where the goal is to erase the data samples of a specific \emph{target client} from the trained model.
Support for client-level unlearning is essential to adhere to privacy regulations such as the right to be forgotten~\cite{gdpr2018}.
We remark that \fump cannot perform client-level unlearning as it is specifically designed for class-level unlearning.
We report evaluations on the \cifar dataset using two different data distributions: Non-IID (with $\alpha=0.1$) and IID (uniform distribution).
The target unlearning client is randomly selected.

\Cref{tab:client_unlearning_different_distribution} shows the performance of \sys against baselines after the unlearning and recovery stages terminate.
We first discuss the non-IID scenario.
We observe that \sys achieves $11.57\%$ on the F-Set, remaining close to the oracle ($10.48\%$) while the baselines achieve $9.58-19.72\%$.
Note that the F-Set accuracies are higher than with class-level unlearning (see~\Cref{Tab:combined_accuracy_computation_comparison}).
This is because even though we unlearned the data samples of a particular client, some features associated with the data of the target client might still be embedded in the model's knowledge through other clients.
Therefore, even after unlearning, some forgotten samples are correctly classified.
Concerning the accuracy on the R-Set, \Cref{tab:client_unlearning_different_distribution} shows that \sys ($70.89\%$) remains very competitive with the baselines ($69.85-72.63\%$) while the oracle achieves the highest ($73.69\%$).
These results are consistent with the accuracies obtained for class-level unlearning on non-IID data (\Cref{Tab:combined_accuracy_computation_comparison}).


In IID settings, we observe even higher accuracies on the \fset after unlearning and recovery. 
For example, \retrainor achieves $70.81\%$, while \sys reaches $68.59\%$. 
This is expected since, with an IID distribution, each client holds a similar type and quantity of data.
Therefore, when we unlearn the target client, much of its contributed knowledge is still represented by the remaining data (R-Set) in the system, and the departure of the target client will barely impact the model performance.
We also explored the effect of different $\alpha$ values ($\alpha={1, 10}$), alongside the reported $\alpha=0.1$ and IID cases. 
Our findings were consistent with previous observations \ie as heterogeneity decreases (larger $\alpha$), the impact of forgetting on final accuracy diminishes.
Nevertheless, \sys remains competitive with the baselines and offers substantial computational efficiency through synthetic data.



\subsection{Performance of Relearning}
\label{subsec:relearning}

This section provides additional accuracy results for single-class unlearning requests with two datasets and a network size of $20$ clients. 
Complementing the previously presented \cifar results with $10$ clients from~\Cref{Tab:combined_accuracy_computation_comparison}, we include the remaining combinations in the left (\cifar with $20$ clients) and the right (\mnist with $20$ clients) partitions of~\Cref{Tab:accuracy_comparison_various_dataset_scale}.
All these experiments follow the same setup as the results shown in~\Cref{Tab:combined_accuracy_computation_comparison}.
We further attach the results of the relearning stage in each table partition to show the effectiveness of different methods in relearning the eliminated class.
The approach used in the relearning stage is the same for different baselines: we adopt traditional SGD-based model training to update the unlearned model with the \forgetset $ D_f $.
Note that \sys still uses the synthetic data in the relearning stage while other baselines use the original data; \sys thus can still keep its superior computation efficiency.

\begin{table}[t]
	\centering
	\caption{The accuracy of \sys and other baselines for \textit{client-level} unlearning on CIFAR-10 (20 clients), with non-IID ($\alpha=0.1$) and IID data distributions.
	We report the performance after unlearning and recovery.}
	\scalebox{1}{
		\begin{tabular}{c|cc|cc}
			\toprule
			\textbf{Distribution} &  \multicolumn{2}{c}{Non-IID ($\alpha=0.1$)}  & \multicolumn{2}{c}{IID}\\
			\cmidrule(r){2-3} \cmidrule(r){4-5}
			\textbf{\ac{FU} approach} & \fset & \multicolumn{1}{c}{R-Set} & \fset & \multicolumn{1}{c}{R-Set} \\
			\cmidrule(r){1-1} \cmidrule(r){2-3} \cmidrule(r){4-5}  
			\retrainor  & $10.48\%$ & $73.69\%$ & $70.81\%$ & $71.64\%$ \\
			\federaser & $16.57\%$ & $69.85\%$ & $ 65.29\%$ & $66.04\%$ \\
			\su & $19.72\%$ & $70.25\%$ & $70.63\%$ & $71.28\%$ \\
			\sgaor & $9.58\%$ & $72.63\%$  & $69.32\%$ & $70.25\%$ \\
			\sys & $11.57\%$  & $70.89\%$  & $68.59\%$ & $68.48\%$ \\
			\bottomrule
	\end{tabular}}
	\label{tab:client_unlearning_different_distribution}
\end{table}

We observe that the performance after unlearning and recovery stages follows a similar trend as \Cref{Tab:combined_accuracy_computation_comparison}.
In particular, almost all algorithms achieve low accuracy on F-Set, similar to that of \retrainor, demonstrating forgetting of the target class on both datasets. 
Concerning the R-Set accuracy on \cifar dataset, \sys ($65.78\%$) remains competitive with the baselines ($67.38\%-70.04\%$) while the oracle achieves $71.48\%$.
This gap is much lower on the \mnist dataset, where \sys achieves $94.26\%$ while the baselines and the oracle obtain $93.52\%-95.63\%$.

\begin{table*}[t]
	\centering
	\caption{The accuracies on the \fset and R-Set after the unlearning + recovery and \textit{relearning} stages, for the \cifar and \mnist datasets with $20$ clients and $\alpha = 0.1$. 
		In each scenario, the goal is to match the performance of \retrainor on both the \fset and the \rset.
	}
	\scalebox{1}{
		\begin{tabular}{c|cc|cc|cc|cc}
			\toprule
			\textbf{Distribution}   & \multicolumn{4}{c}{CIFAR-10 (20 clients, $\alpha=0.1$)}  & \multicolumn{4}{|c}{MNIST (20 clients, $\alpha=0.1$)} \\
			\cmidrule(r){2-5} \cmidrule(r){6-9} 
			\textbf{Stage} & \multicolumn{2}{c}{Unlearning + Recovery}   & \multicolumn{2}{c}{Relearning} & \multicolumn{2}{|c}{Unlearning + Recovery}   & \multicolumn{2}{c}{Relearning} \\
			\cmidrule(r){2-3} \cmidrule(r){4-5} \cmidrule(r){6-7}  \cmidrule(r){8-9}
			\textbf{\ac{FU} approach} & \fset & \multicolumn{1}{c}{R-Set} & \fset & \multicolumn{1}{c}{R-Set}  & \multicolumn{1}{|c}{\fset}  & \multicolumn{1}{c}{R-Set} & \fset & \multicolumn{1}{c}{R-Set}\\
			\cmidrule(r){1-1} \cmidrule(r){2-3} \cmidrule(r){4-5} \cmidrule(r){6-7}  \cmidrule(r){8-9}
			\retrainor  & $0.68\%$ & $71.48\%$ & $78.65\%$ & $71.83\%$ & $0.47\%$ & $95.63\%$ & $96.82\%$ & $95.74\%$ \\
			\federaser & $0.22\%$ & $67.38\%$ & $70.48\%$ & $68.22\%$ & $0.23\%$ & $93.52\%$ & $95.86\%$ & $95.43\%$ \\
			\sgaor & $0.71\%$ & $70.04\%$ & $75.83\%$ & $69.75\%$ & $0.51\%$ & $95.03\%$ & $96.28\%$ & $95.18\%$ \\
			\fump & $0.59\%$ & $69.82\%$ & --- & --- & $0.31\%$ & $94.83\%$ & --- & ---  \\
			\sys & $0.69\%$ & $65.78\%$ & $74.39\%$ & $66.21\%$  & $0.44\%$ & $94.26\%$ & $96.37\%$ & $94.58\%$ \\
			\bottomrule
	\end{tabular}}
	\label{Tab:accuracy_comparison_various_dataset_scale}
\end{table*}

\Cref{Tab:accuracy_comparison_various_dataset_scale} also reports the accuracy on the \fset and R-Set after relearning.
Ideally, we want these accuracies to be high since we attempt to restore the model to the state before unlearning.
\Cref{Tab:accuracy_comparison_various_dataset_scale} shows that all evaluated \ac{FU} approaches successfully relearn the previously eliminated knowledge again.
On the \mnist dataset, \sys achieves an accuracy of $96.37\%$ on the F-Set and $94.58\%$ on the R-Set, almost matching \retrainor ($96.82\%$ and $95.74\%$).
At the same time, \sys keeps its superiority in computation efficiency since the relearning stage uses the compact synthetic dataset ($66.7 \times$ faster than \retrainor and $47.29 \times$ than \sgaor).
We are unable to relearn using \fump.
This is because model pruning irreversibly modifies the model structure, as described in \Cref{subsec:existing_algos}.
In conclusion, \sys demonstrates high versatility in not only efficiently unlearning target classes but also relearning at low computation cost.

\subsection{Computational Overhead of \sys}
\label{subsec:exp_compute_overhead}
\sys generates synthetic samples during the \ac{FL} training process, which incurs computational overhead.
While we are able to re-use the gradients on original data ($\nabla\ml^{\md_i}$) computed by \ac{FL}, we are still required to compute gradients on the synthetic data ($\nabla\ml^{\ms_i}$) and update the synthetic samples (line 14 and 15 in~\Cref{algo:quickdrop_dd}) during \ac{FL} training.
\Cref{Tab:dd_compute_overhead} shows the total and \ac{DD} compute time for the 3 different datasets used across our experiments.
The total time corresponds to the time required for \ac{FL} training in \sys, which includes \ac{DD} time.
The \ac{DD} time corresponds to the time spent in executing line 14 and line 15 in \Cref{algo:quickdrop_dd}.
We also compute the overhead of \ac{DD} as percentage of the total compute time.
\Cref{Tab:dd_compute_overhead} shows that the compute overhead of \ac{DD} ranges around $50\%$, \ie it doubles the \ac{FL} training time.  
Although \sys slows down \ac{FL} training, this initial investment is necessary to unlock significant efficiency gains in downstream unlearning as we showed previously.

\begin{table}[b]
	\centering
	\caption{The total and \ac{DD} compute time, and the overhead of \ac{DD}, for all three datasets used in our experiments.
	}
	\scalebox{1}{
		\begin{tabular}{c|c|c|c}
			\toprule
			\textbf{Dataset} & \makecell{\textbf{Total Compute}\\ \textbf{Time (s)}} & \makecell{\textbf{\ac{DD} Compute}\\ \textbf{Time (s)}} & \textbf{Overhead} \\ \hline
			\mnist & 4735 & 2557 & 54\% \\
			\cifar & 5360 & 2948 & 55\% \\
			\svhn & 9079 & 4204 & 46.3\% \\
			\bottomrule
	\end{tabular}}
	\label{Tab:dd_compute_overhead}
\end{table}

\section{Discussion}
\label{sec:discussion}


We now discuss certain aspects of our evaluation and algorithmic aspects of \sys that were not addressed before.
We also discuss two limitations of \sys in~\Cref{subsec:limitations}.

\textbf{Unlearning Metric.}
First, we highlight that developing new metrics for unlearning is an active area of research~\cite{triantafillou2024we}.
While accuracy might not fully capture true unlearning, our approach aligns with \ac{SOTA} \ac{FU} work that reports unlearning via the accuracy metric~\cite{wang2022federated,gao2022verifi,liu2021federaser}. 
Additionally, we reported the \ac{MIA} performance to provide another metric for assessing unlearning.

\textbf{Privacy of \sys.}
We point out that the synthetically generated data in \sys does not leave the client's device.
Clients only exchange model parameters with the server during the unlearning and recovery phases.
Hence, the privacy guarantees for \sys remain the same as standard \ac{FL}.

\textbf{Partial Client Participation.}
In \ac{FL} with many clients, \ac{FL} algorithms usually have the parameter server choose a subset of clients to participate in model training during each round~\cite{charles2021large}.
\sys supports such partial client participation during both \ac{FL} training and unlearning.
Our \svhn experiments use a $10\%$ participation rate for \ac{FL} training.
Even though the clients participate and consequently synthesize data only in a few rounds, the quality of unlearning is good.
This is because data synthesis aims to eliminate only \textit{contributed} knowledge, \ie, corresponding to the round in which the client participated.
Thus, this goal is met even with partial participation.
Partial participation during unlearning and recovery is directly applicable in \sys.

\textbf{Frequency of Unlearning Requests.}
The computational benefits of \sys are closely tied to the frequency of unlearning requests, \ie, the benefits in compute time are effectively realized as there are more unlearning requests in the system.
These requests can arise from individual clients exercising their right to be forgotten or organisations needing to remove outdated or erroneous data.
\sys requires an upfront investment in generating the synthetic dataset, but these compute costs are then amortized over the subsequent unlearning requests.
Existing research on machine unlearning studies performance with hundreds of requests~\cite{bourtoule2021machine} and highlights the importance of efficiently managing multiple requests~\cite{zhang2022prompt}.
Recent work also proposes a streaming unlearning setting involving a sequence of data removal requests~\cite{gupta2021adaptive}.
We thus anticipate a similar or even greater demand for efficiently managing multiple unlearning requests when performing federated unlearning.


\subsection{Limitations}
\label{subsec:limitations}
Finally, we discuss two limitations of our approach.

\textbf{Sample-level Unlearning.}
In its current form, \sys supports only class-level and client-level unlearning.
These two levels of unlearning already cover many applications of machine unlearning in federated setups.
One might want to perform sample-level unlearning, where the goal is to unlearn a subset of data samples of a particular client.
In \sys, clients locally generate class-wise synthetic data. 
Hence, one way to adapt \sys to enable sample-level unlearning is to consider subsets of data within each class.
One can generate synthetic samples for each subset and then unlearn at the granularity of these subsets.
We consider this challenge beyond the scope of current work and leave the exploration of these ideas for future work. 

\textbf{Compute and Storage Overhead.}
\sys prolongs the training time for the original \ac{FL} training, and requires additional storage for the synthetic datasets.
We have evaluated \sys's compute overhead in~\Cref{subsec:exp_compute_overhead}, showing that this overhead is between 46.3\% and 55\% in terms of prolonged training time on our evaluated datasets.
The storage overhead of \sys, determined by the scale parameter $s$, is $ \frac{1}{s} $ of the local training dataset size per client.
In~\Cref{subsec:exp_scale_factor}, we have experimentally shown  that decreasing the scale increases the size of the synthetic dataset but allows for more accurate unlearning.
We find $ s = 100 $ to yield a reasonable balance between efficiency and effectiveness and use this value across all our experiments.
This results in $1\%$ storage overhead for \sys.

\section{Related Work}
\label{sec:related_work}


We now discuss related work in the domain of machine and federated unlearning (\Cref{subsec:related_work_mu_fu}) and dataset distillation (\Cref{subsec:related_work_dd}).

\subsection{Machine and Federated Unlearning}
\label{subsec:related_work_mu_fu}
Following the introduction of \ac{MU} in \cite{cao2015towards}, several algorithms have been proposed~\cite{du2019lifelong, ginart2019making, guo2019certified, golatkar2020eternal, golatkar2020forgetting, bourtoule2021machine}.
These works focus mainly on unlearning knowledge from simple classification models, \eg, for logistic regression, but are unsuitable for more complex models, \eg, deep neural networks.
Some algorithms have other restrictions and can only be applied to specific model architectures or scenarios, \eg, \cite{brophy2021machine} only fits random forests model and \cite{nguyen2020variational} is only for Bayesian learning.
In traditional machine learning scenarios, all the local data of clients will be uploaded to the server for centralized management, so the server has a high level of flexibility to conduct arbitrary operations on all data.
Therefore, various unlearning techniques (\eg, ensemble learning for data splitting \cite{bourtoule2021machine} or gradient amnesia of arbitrary batch \cite{graves2021amnesiac}) are designed to operate in settings where training data is readily available.
Such operations at data-level are not possible in \ac{FU}, thus \ac{FU} is more difficult than \ac{MU} as outlined in ~\Cref{subsec:mu_and_fu}.
Besides the \ac{FU} methods covered in \Cref{subsec:existing_algos}, other recent approaches include~\cite{halimi2022federated,yuan2023federated}.
These works are not adopted as baselines either because they employ a similar unlearning technique as one of our baselines (\eg \cite{halimi2022federated} is built upon \sga) or are tailored to specific scenarios (\eg, \cite{yuan2023federated} focuses on recommendation systems).
Finally, among the very few theoretical works on \ac{FU}, \cite{10.14778/3641204.3641220} tackle both communication efficiency and provable exact unlearning by leveraging total variation stability.
We refer the reader to \cite{10.1145/3679014} for a more comprehensive overview of \ac{FU}.



\subsection{Dataset Distillation}
\label{subsec:related_work_dd}
Standard \ac{DD} aims to replace a large training dataset with a significantly smaller one that can achieve the same generalization performance as the original training data~\cite{wang2018dataset}.
DD can potentially speed up downstream tasks such as continual learning~\cite{hadsell2020embracing} and neural architecture search~\cite{ren2021comprehensive}.
\ac{DD} also has been previously leveraged in one-shot \ac{FL} to significantly reduce communication cost compared to multi-round \ac{FL}~\cite{10191879,zhou2020distilled}. 
Early DD approaches are based on core-set selection \ie identifying a subset of influential samples during training~\cite{bachem2017practical,sener2018active}.
Another class of algorithms synthesizes a set of new samples from the original dataset.
The approach described in~\cite{zhao2021dataset} is to match the gradients of a model trained on the original and synthetic data.
Follow-up work has introduced distillation techniques based on trajectory gradient matching~\cite{cazenavette2022dataset}, differential data augmentation functions~\cite{zhao2021dataset_other}, distribution matching~\cite{zhao2023dataset} and feature alignment~\cite{wang2022cafe}.
While existing \ac{DD} methods achieve leading performance, synthesizing samples targeted at generalization is highly compute intensive~\cite{zhang2023accelerating}.
\sys, in contrast, formulates \ac{DD} for unlearning and synthesizes samples at lower computational overhead.

\section{Conclusion}
\label{sec:conclusion}

We introduced \sys, a novel and efficient federated unlearning method in which clients generate and use synthetic datasets for unlearning.
These synthetic datasets are a compact representation of the gradient information generated during training.
To unlearn specific samples, clients execute stochastic gradient ascent (SGA) with synthetic dataset instead of the original training data.
Recovery on remaining samples also takes place via synthetic datasets.
Empirical evaluations using three standard datasets and \ac{SOTA} baselines confirm the effectiveness and efficiency of \sys, demonstrating a remarkable acceleration in the unlearning process compared to existing federated unlearning approaches.


\begin{acks}
This work has been funded by the Swiss National Science Foundation, under the project ``FRIDAY: Frugal, Privacy-Aware and Practical Decentralized Learning'', SNSF proposal No. 10.001.796. 
This work was also supported by fundings from the Key-Area Research and Development Program of Guangdong Province (No. 2021B0101400003), Hong Kong RGC Research Impact Fund (No. R5060-19, No. R5034-18, No. R5011-23), Areas of Excellence Scheme (AoE/E-601/22-R), General Research Fund (No. 152203/20E, 152244/ 21E, 152169/22E, 152228/23E), Collaborative Research Fund (No. 8730102).
\end{acks}

\bibliographystyle{plain}
\bibliography{references}

\begin{thebibliography}{10}

\bibitem{bachem2017practical}
Olivier Bachem, Mario Lucic, and Andreas Krause.
\newblock Practical coreset constructions for machine learning.
\newblock {\em arXiv preprint arXiv:1703.06476}, 2017.

\bibitem{bellet2018personalized}
Aur{\'e}lien Bellet, Rachid Guerraoui, Mahsa Taziki, and Marc Tommasi.
\newblock Personalized and private peer-to-peer machine learning.
\newblock In {\em ICAIS}, pages 473--481. PMLR, 2018.

\bibitem{bourtoule2021machine}
Lucas Bourtoule, Varun Chandrasekaran, Christopher~A Choquette-Choo, Hengrui Jia, Adelin Travers, Baiwu Zhang, David Lie, and Nicolas Papernot.
\newblock Machine unlearning.
\newblock In {\em 2021 IEEE Symposium on Security and Privacy (SP)}, pages 141--159. IEEE, 2021.

\bibitem{brophy2021machine}
Jonathan Brophy and Daniel Lowd.
\newblock Machine unlearning for random forests.
\newblock In {\em International Conference on Machine Learning}, pages 1092--1104. PMLR, 2021.

\bibitem{ccpa2018}
{California State Legislature}.
\newblock California consumer privacy act of 2018.
\newblock California Legislative Information, 2018.
\newblock Available online: \url{https://leginfo.legislature.ca.gov/faces/billTextClient.xhtml?bill_id=201720180AB375}.

\bibitem{cao2015towards}
Yinzhi Cao and Junfeng Yang.
\newblock Towards making systems forget with machine unlearning.
\newblock In {\em 2015 IEEE symposium on security and privacy}, pages 463--480. IEEE, 2015.

\bibitem{cazenavette2022dataset}
George Cazenavette, Tongzhou Wang, Antonio Torralba, Alexei~A Efros, and Jun-Yan Zhu.
\newblock Dataset distillation by matching training trajectories.
\newblock In {\em Proceedings of the IEEE/CVF Conference on Computer Vision and Pattern Recognition}, pages 4750--4759, 2022.

\bibitem{charles2021large}
Zachary Charles, Zachary Garrett, Zhouyuan Huo, Sergei Shmulyian, and Virginia Smith.
\newblock On large-cohort training for federated learning.
\newblock {\em Advances in neural information processing systems}, 34:20461--20475, 2021.

\bibitem{chen2021machine}
Min Chen, Zhikun Zhang, Tianhao Wang, Michael Backes, Mathias Humbert, and Yang Zhang.
\newblock When machine unlearning jeopardizes privacy.
\newblock In {\em Proceedings of the 2021 ACM SIGSAC conference on computer and communications security}, pages 896--911, 2021.

\bibitem{chundawat2023zero}
Vikram~S Chundawat, Ayush~K Tarun, Murari Mandal, and Mohan Kankanhalli.
\newblock Zero-shot machine unlearning.
\newblock {\em IEEE Transactions on Information Forensics and Security}, 2023.

\bibitem{du2019lifelong}
Min Du, Zhi Chen, Chang Liu, Rajvardhan Oak, and Dawn Song.
\newblock Lifelong anomaly detection through unlearning.
\newblock In {\em Proceedings of the 2019 ACM SIGSAC conference on computer and communications security}, pages 1283--1297, 2019.

\bibitem{gdpr2018}
{European Union}.
\newblock Regulation (eu) 2016/679 of the european parliament and of the council of 27 april 2016 on the protection of natural persons with regard to the processing of personal data and on the free movement of such data, and repealing directive 95/46/ec (general data protection regulation).
\newblock Official Journal of the European Union, 2018.
\newblock OJ L 119, 4.5.2016, p. 1–88.

\bibitem{gao2022verifi}
Xiangshan Gao, Xingjun Ma, Jingyi Wang, Youcheng Sun, Bo~Li, Shouling Ji, Peng Cheng, and Jiming Chen.
\newblock Verifi: Towards verifiable federated unlearning.
\newblock {\em arXiv preprint arXiv:2205.12709}, 2022.

\bibitem{gidaris2018dynamic}
Spyros Gidaris and Nikos Komodakis.
\newblock Dynamic few-shot visual learning without forgetting.
\newblock In {\em CVPR}, pages 4367--4375, 2018.

\bibitem{ginart2019making}
Antonio Ginart, Melody Guan, Gregory Valiant, and James~Y Zou.
\newblock Making ai forget you: Data deletion in machine learning.
\newblock {\em Advances in neural information processing systems}, 32, 2019.

\bibitem{golatkar2021mixed}
Aditya Golatkar, Alessandro Achille, Avinash Ravichandran, Marzia Polito, and Stefano Soatto.
\newblock Mixed-privacy forgetting in deep networks.
\newblock In {\em Proceedings of the IEEE/CVF Conference on Computer Vision and Pattern Recognition}, pages 792--801, 2021.

\bibitem{golatkar2020eternal}
Aditya Golatkar, Alessandro Achille, and Stefano Soatto.
\newblock Eternal sunshine of the spotless net: Selective forgetting in deep networks.
\newblock In {\em CVPR}, pages 9304--9312, 2020.

\bibitem{golatkar2020forgetting}
Aditya Golatkar, Alessandro Achille, and Stefano Soatto.
\newblock Forgetting outside the box: Scrubbing deep networks of information accessible from input-output observations.
\newblock In {\em Computer Vision--ECCV 2020: 16th European Conference, Glasgow, UK, August 23--28, 2020, Proceedings, Part XXIX 16}, pages 383--398. Springer, 2020.

\bibitem{graves2021amnesiac}
Laura Graves, Vineel Nagisetty, and Vijay Ganesh.
\newblock Amnesiac machine learning.
\newblock In {\em Proceedings of the AAAI Conference on Artificial Intelligence}, volume~35, pages 11516--11524, 2021.

\bibitem{guo2019certified}
Chuan Guo, Tom Goldstein, Awni Hannun, and Laurens Van Der~Maaten.
\newblock Certified data removal from machine learning models.
\newblock {\em arXiv preprint arXiv:1911.03030}, 2019.

\bibitem{gupta2021adaptive}
Varun Gupta, Christopher Jung, Seth Neel, Aaron Roth, Saeed Sharifi-Malvajerdi, and Chris Waites.
\newblock Adaptive machine unlearning.
\newblock {\em Advances in Neural Information Processing Systems}, 34:16319--16330, 2021.

\bibitem{hadsell2020embracing}
Raia Hadsell, Dushyant Rao, Andrei~A Rusu, and Razvan Pascanu.
\newblock Embracing change: Continual learning in deep neural networks.
\newblock {\em Trends in cognitive sciences}, 24(12):1028--1040, 2020.

\bibitem{halimi2022federated}
Anisa Halimi, Swanand Kadhe, Ambrish Rawat, and Nathalie Baracaldo.
\newblock Federated unlearning: How to efficiently erase a client in fl?
\newblock {\em arXiv preprint arXiv:2207.05521}, 2022.

\bibitem{hsu2019measuring}
Tzu-Ming~Harry Hsu, Hang Qi, and Matthew Brown.
\newblock Measuring the effects of non-identical data distribution for federated visual classification.
\newblock {\em arXiv preprint arXiv:1909.06335}, 2019.

\bibitem{krizhevsky2009learning}
Alex Krizhevsky, Geoffrey Hinton, et~al.
\newblock Learning multiple layers of features from tiny images.
\newblock 2009.

\bibitem{lecun1998mnist}
Yann LeCun.
\newblock The mnist database of handwritten digits.
\newblock {\em http://yann. lecun. com/exdb/mnist/}, 1998.

\bibitem{liu2021federaser}
Gaoyang Liu, Xiaoqiang Ma, Yang Yang, Chen Wang, and Jiangchuan Liu.
\newblock Federaser: Enabling efficient client-level data removal from federated learning models.
\newblock In {\em 2021 IEEE/ACM 29th International Symposium on Quality of Service (IWQOS)}, pages 1--10. IEEE, 2021.

\bibitem{10.1145/3679014}
Ziyao Liu, Yu~Jiang, Jiyuan Shen, Minyi Peng, Kwok-Yan Lam, Xingliang Yuan, and Xiaoning Liu.
\newblock A survey on federated unlearning: Challenges, methods, and future directions.
\newblock {\em ACM Comput. Surv.}, 57(1), October 2024.

\bibitem{mahdavinejad2018machine}
Mohammad~Saeid Mahdavinejad, Mohammadreza Rezvan, Mohammadamin Barekatain, Peyman Adibi, Payam Barnaghi, and Amit~P Sheth.
\newblock Machine learning for internet of things data analysis: A survey.
\newblock {\em Digital Communications and Networks}, 4(3):161--175, 2018.

\bibitem{mcmahan2017communication}
Brendan McMahan, Eider Moore, Daniel Ramage, Seth Hampson, and Blaise~Aguera y~Arcas.
\newblock Communication-efficient learning of deep networks from decentralized data.
\newblock In {\em AISTATS}, pages 1273--1282. PMLR, 2017.

\bibitem{netzer2011reading}
Yuval Netzer, Tao Wang, Adam Coates, Alessandro Bissacco, Bo~Wu, and Andrew~Y Ng.
\newblock Reading digits in natural images with unsupervised feature learning.
\newblock 2011.

\bibitem{nguyen2020variational}
Quoc~Phong Nguyen, Bryan Kian~Hsiang Low, and Patrick Jaillet.
\newblock Variational bayesian unlearning.
\newblock {\em Advances in Neural Information Processing Systems}, 33:16025--16036, 2020.

\bibitem{ren2021comprehensive}
Pengzhen Ren, Yun Xiao, Xiaojun Chang, Po-Yao Huang, Zhihui Li, Xiaojiang Chen, and Xin Wang.
\newblock A comprehensive survey of neural architecture search: Challenges and solutions.
\newblock {\em ACM Computing Surveys (CSUR)}, 54(4):1--34, 2021.

\bibitem{romandini2024federated}
Nicol{\`o} Romandini, Alessio Mora, Carlo Mazzocca, Rebecca Montanari, and Paolo Bellavista.
\newblock Federated unlearning: A survey on methods, design guidelines, and evaluation metrics.
\newblock {\em arXiv preprint arXiv:2401.05146}, 2024.

\bibitem{sener2018active}
Ozan Sener and Silvio Savarese.
\newblock Active learning for convolutional neural networks: A core-set approach.
\newblock In {\em International Conference on Learning Representations}, 2018.

\bibitem{10191879}
Rui Song, Dai Liu, Dave~Zhenyu Chen, Andreas Festag, Carsten Trinitis, Martin Schulz, and Alois Knoll.
\newblock Federated learning via decentralized dataset distillation in resource-constrained edge environments.
\newblock In {\em 2023 International Joint Conference on Neural Networks (IJCNN)}, pages 1--10, 2023.

\bibitem{10.14778/3641204.3641220}
Youming Tao, Cheng-Long Wang, Miao Pan, Dongxiao Yu, Xiuzhen Cheng, and Di~Wang.
\newblock Communication efficient and provable federated unlearning.
\newblock {\em Proc. VLDB Endow.}, 17(5):1119–1131, May 2024.

\bibitem{triantafillou2024we}
Eleni Triantafillou, Peter Kairouz, Fabian Pedregosa, Jamie Hayes, Meghdad Kurmanji, Kairan Zhao, Vincent Dumoulin, Julio~Jacques Junior, Ioannis Mitliagkas, Jun Wan, et~al.
\newblock Are we making progress in unlearning? findings from the first neurips unlearning competition.
\newblock {\em arXiv preprint arXiv:2406.09073}, 2024.

\bibitem{wang2022federated}
Junxiao Wang, Song Guo, Xin Xie, and Heng Qi.
\newblock Federated unlearning via class-discriminative pruning.
\newblock In {\em Proceedings of the ACM Web Conference 2022}, pages 622--632, 2022.

\bibitem{wang2022cafe}
Kai Wang, Bo~Zhao, Xiangyu Peng, Zheng Zhu, Shuo Yang, Shuo Wang, Guan Huang, Hakan Bilen, Xinchao Wang, and Yang You.
\newblock Cafe: Learning to condense dataset by aligning features.
\newblock In {\em CVPR}, pages 12196--12205, 2022.

\bibitem{wang2018dataset}
Tongzhou Wang, Jun-Yan Zhu, Antonio Torralba, and Alexei~A Efros.
\newblock Dataset distillation.
\newblock {\em arXiv preprint arXiv:1811.10959}, 2018.

\bibitem{wu2022federated}
Leijie Wu, Song Guo, Junxiao Wang, Zicong Hong, Jie Zhang, and Yaohong Ding.
\newblock Federated unlearning: Guarantee the right of clients to forget.
\newblock {\em IEEE Network}, 36(5):129--135, 2022.

\bibitem{yang2019federated}
Qiang Yang, Yang Liu, Tianjian Chen, and Yongxin Tong.
\newblock Federated machine learning: Concept and applications.
\newblock {\em ACM TIST}, 10(2):1--19, 2019.

\bibitem{yuan2023federated}
Wei Yuan, Hongzhi Yin, Fangzhao Wu, Shijie Zhang, Tieke He, and Hao Wang.
\newblock Federated unlearning for on-device recommendation.
\newblock In {\em Proceedings of the Sixteenth ACM International Conference on Web Search and Data Mining}, pages 393--401, 2023.

\bibitem{zhang2023review}
Haibo Zhang, Toru Nakamura, Takamasa Isohara, and Kouichi Sakurai.
\newblock A review on machine unlearning.
\newblock {\em SN Computer Science}, 4(4):337, 2023.

\bibitem{zhang2023accelerating}
Lei Zhang, Jie Zhang, Bowen Lei, Subhabrata Mukherjee, Xiang Pan, Bo~Zhao, Caiwen Ding, Yao Li, and Dongkuan Xu.
\newblock Accelerating dataset distillation via model augmentation.
\newblock In {\em Proceedings of the IEEE/CVF Conference on Computer Vision and Pattern Recognition}, pages 11950--11959, 2023.

\bibitem{zhang2022prompt}
Zijie Zhang, Yang Zhou, Xin Zhao, Tianshi Che, and Lingjuan Lyu.
\newblock Prompt certified machine unlearning with randomized gradient smoothing and quantization.
\newblock {\em Advances in Neural Information Processing Systems}, 35:13433--13455, 2022.

\bibitem{zhao2021dataset_other}
Bo~Zhao and Hakan Bilen.
\newblock Dataset condensation with differentiable siamese augmentation.
\newblock In {\em International Conference on Machine Learning}, pages 12674--12685. PMLR, 2021.

\bibitem{zhao2023dataset}
Bo~Zhao and Hakan Bilen.
\newblock Dataset condensation with distribution matching.
\newblock In {\em Proceedings of the IEEE/CVF Winter Conference on Applications of Computer Vision}, pages 6514--6523, 2023.

\bibitem{zhao2021dataset}
Bo~Zhao, Konda~Reddy Mopuri, and Hakan Bilen.
\newblock Dataset condensation with gradient matching.
\newblock In {\em International Conference on Learning Representations}, 2021.

\bibitem{zhou2020distilled}
Yanlin Zhou, George Pu, Xiyao Ma, Xiaolin Li, and Dapeng Wu.
\newblock Distilled one-shot federated learning.
\newblock {\em arXiv preprint arXiv:2009.07999}, 2020.

\bibitem{zhu2021federated}
Hangyu Zhu, Jinjin Xu, Shiqing Liu, and Yaochu Jin.
\newblock Federated learning on non-iid data: A survey.
\newblock {\em Neurocomputing}, 465:371--390, 2021.

\bibitem{zhu2021data}
Zhuangdi Zhu, Junyuan Hong, and Jiayu Zhou.
\newblock Data-free knowledge distillation for heterogeneous federated learning.
\newblock In {\em ICML}, pages 12878--12889. PMLR, 2021.

\end{thebibliography}

\end{document}